\documentclass[conference]{IEEEtran}
\usepackage{amsmath,amssymb,amsfonts}

\usepackage[ruled,norelsize,linesnumbered]{algorithm2e}
\usepackage{color}
\usepackage{graphicx}
\usepackage{microtype}

\newcommand{\Real}{\mathbb{R}}
\newcommand{\Natural}{\mathbb{N}}
\newcommand{\reject}{\ell^{\times}}
\newcommand{\Cov}{\mathrm{Cov}}
\newcommand{\Mean}{\mathrm{Mean}}

\newtheorem{prob}{Problem}
\newtheorem{thm}{Theorem}
\newtheorem{lem}{Lemma}

\ifCLASSINFOpdf
\else
\fi
\begin{document}
%
\title{Weakly Supervised Learners for Correction of AI Errors with Provable Performance Guarantees}



\author{\IEEEauthorblockN{Ivan Y. Tyukin}
\IEEEauthorblockA{Department of Mathematics\\
King's College London\\
Email: ivan.tyukin@kcl.ac.uk}
\and
\IEEEauthorblockN{Tatiana Tyukina}
\IEEEauthorblockA{School of Computing\\ and Mathematical Sciences\\
University of Leicester\\
Email: tt51@le.ac.uk}
\and
\IEEEauthorblockN{Daniel van Helden}
\IEEEauthorblockA{School of Archaeology\\ and Ancient History\\
University of Leicester\\
Email: dpvh2@le.ac.uk}
\and
\IEEEauthorblockN{Zedong Zheng}
\IEEEauthorblockA{School of Computing\\ and Mathematical Sciences\\
University of Leicester\\
Email: zz288@le.ac.uk}
\and
\IEEEauthorblockN{Evgeny M. Mirkes}
\IEEEauthorblockA{School of Computing\\ and Mathematical Sciences\\
University of Leicester\\
Email: em322@le.ac.uk}
\and
\IEEEauthorblockN{Oliver J. Sutton}
\IEEEauthorblockA{Department of Mathematics\\
King's College London\\
Email: oliver.sutton@kcl.ac.uk}
\and
\IEEEauthorblockN{Qinghua Zhou}
\IEEEauthorblockA{Department of Mathematics\\
King's College London\\
Email: qinghua.zhou@kcl.ac.uk}
\and
\IEEEauthorblockN{Alexander N. Gorban}
\IEEEauthorblockA{School of Computing\\ and Mathematical Sciences\\
University of Leicester\\
Email: ag153@le.ac.uk}
\and
\centering
\IEEEauthorblockN{\hspace{70mm} Penelope Allison}
\IEEEauthorblockA{\hspace{70mm} School of Archaeology\\ \hspace{70mm} and Ancient History\\
\hspace{70mm} University of Leicester\\
\hspace{70mm} Email: pma9@le.ac.uk}
}


%


\maketitle

\begin{abstract}
We present a new methodology for handling AI errors by introducing weakly supervised AI error correctors with \emph{a priori} performance guarantees.
These AI correctors are auxiliary maps whose role is to moderate the decisions of some previously constructed underlying classifier by either approving or rejecting its decisions.
The rejection of a decision can be used as a signal to suggest abstaining from making a decision.
A key technical focus of the work is in providing performance guarantees for these new AI correctors through bounds on the probabilities of incorrect decisions.
These bounds are distribution agnostic and do not rely on assumptions on the data dimension.
Our empirical example illustrates how the framework can be applied to improve the performance of an image classifier in a challenging real-world task where training data are scarce.
\end{abstract}


%
\IEEEpeerreviewmaketitle

\section{Introduction}

Over the last few decades, AI systems have undergone a significant transformation, from mere proof-of-principle solutions confined to the realms of lab-based research and studies to real-life deployment with applications across many sectors. In many areas, such as image classification, chess, and go, these systems demonstrate super-human performance. More recently, with the advent of transformers and large language models, AI systems built from data are now capable of mimicking sustained human-level conversation and are getting close to passing the Turing Test --- the task which, until recently, no machine was able to complete \cite{TuringTest2023}.

Despite these and many other successes,  there are several fundamental open questions surrounding this new class tools and technology. One such issue is the challenge of AI errors, which are inherent in the majority of AI systems built with empirically collected data. Numerous and growing evidence of AI incidents \cite{ai_incidents} indicates that the problem is widespread across sectors. It is also notoriously difficult to handle by mere re-training which is also expensive for large-scale models.

Sources of AI errors are numerous and include unavoidable uncertainties, noise, and imperfections associated with any physical measurement. They could be facilitated by AI instability \cite{colbrook2022difficulty}, \cite{Hansen2020}, under-sampling or dark data \cite{dark_data}, and concept drifts \cite{Concept_drift}. Recent works \cite{bastounis2021mathematics,bastounis2023boundaries} showed that there is, unfortunately, an uncountably large class of tasks for which stable and unstable neural network models may co-exist within the same architecture, even with arbitrarily similar weights. The costs of checking both the accuracy and stability of models in these tasks can scale exponentially with input dimension, both in terms of data requirements and computational cost. 

Additionally, the practical application of AI models may be hindered by the foundational assumption that the statistical characteristics of the environment into which the AI system is deployed are identical to those of the data sample the system was built on.
Violating this assumption makes classical generalisation bounds and sample complexity estimates stemming from the principle of empirical risk minimisation \cite{vapnik1991principles} inapplicable in practice.
Moreover, even if  pre- or post-deployment model calibration is used to adapt some aspects of model performance to its intended operational environments (see e.g. \cite{lee2019review}, \cite{wang2023calibration}, \cite{kuleshov2022calibrated}), calibration alone may not be able to improve the accuracy of the final solution \cite{zhang2020effect}.

A promising approach for mitigating the challenge of AI errors
has been suggested in a series of works by Gorban et al.~\cite{gorban2016blessing, gorban2018correction, gorban2019one, gorban2021high}. The solution is to introduce additional devices -- AI correctors -- which could be easily designed and embedded into existing AI, including strong and large models. These devices could fix AI errors as soon as they are detected without the need for expensive retraining of the core model. 

In this work, we present a novel approach to address the challenge of AI errors.
The approach builds on and extends the technology of AI error correctors whose function is to moderate the existing system's decisions \cite{gorban2019one,gorban2018correction}. These correctors are weakly supervised since the amounts of labelled data needed to train these are generally not supposed to be large. In contrast to previous works in this direction, we do not wish to rely on the high intrinsic dimensionality of the data or make any assumptions on the probability distribution the data are sampled from.
Instead, our ambition is to derive rigorous performance guarantees on the corrector's decisions in the most general setting, which is achieved by providing bounds on:
\begin{enumerate}
    \item the conditional probability that the corrector recommends accepting the pre-existing AI system's decision given that the AI's decision is correct
    \item the conditional probability that the corrector recommends rejecting the pre-existing AI system's decision given that the AI's decision is wrong. 
\end{enumerate}

These conditional probabilities are reminiscent of the classical sensitivity and specificity measures and, as such, constitute objective performance characterisations which are more robust than classification accuracy, which is sensitive to class imbalance.
Importantly, the availability of these bounds enables the correctors to be used to produce a recommendation to abstain from making a decision --- a highly relevant option to enhance trust and performance post-training \cite{abstaining_2021}.
However, existing theoretical and practical results in this area require knowledge of the probabilities of errors \cite{chow1970optimum}, the costs of making an error \cite{bartlett2008classification}, or both.
Here we aim to develop tools that could be used as a source of such knowledge, opening a pathway towards applications of the theory developed to date in tasks and problems that could not previously be addressed by existing frameworks.

The paper is organised as follows. 
In Section~\ref{sec:notation} we introduce relevant notation, Section~\ref{sec:Problem} states the problem considered in this work.
Main theoretical results are presented in Section~\ref{sec:Main}. 
Section~\ref{sec:Example} provides an illustrative example of the proposed approach to a challenging classification task where labelled training data is lacking, and Section~\ref{sec:Conclusion} concludes the paper.

\section{Notation}\label{sec:notation}

The following  notational agreements are adopted throughout the paper: $\Natural$ denotes the set of natural numbers, $\Real$ denotes the field of real numbers, and $\Real^n$ denotes the $n$-dimensional real vector space; $({x},{y})=\sum_{k=1}^n x_{k} y_{k}$ is the Euclidean inner product of ${x}\in\Real^n$ and ${y}\in\Real^n$, and $\|{x}\|=\sqrt{({x},{x})}$ is the standard Euclidean norm in $\Real^n$. For a monotone function $f:X\rightarrow Y$, where $X,Y$ are intervals in $\Real$, we shall use $f^\dagger$ denote the pseudo-inverse of $f$: 
\[f^\dagger(y)=\inf\{x\in X \,| \, f(x)\geq y\}.
\]If $\mathcal{S}$ is a set then $|\mathcal{S}|$ is the cardinality of $\mathcal{S}$. For a set $\mathcal{X}$ and a subset $\mathcal{Y} \subset \mathcal{X}$, we use $\boldsymbol{1}_{\mathcal{Y}}:\mathcal{X}\rightarrow \{0,1\}$ to denote the indicator function of $\mathcal{Y}$, such that $\boldsymbol{1}_{\mathcal{Y}}(x)=1$ if $x\in\mathcal{Y}$ and $\boldsymbol{1}_{\mathcal{Y}}(x)=0$ otherwise. If $P_D$ is a probability distribution then $z \sim P_D$ denotes $z$ randomly drawn from the distribution $P_D$.

\section{Problem Formulation}\label{sec:Problem}

Suppose that we have a pre-trained classifier $F$
\begin{equation}\label{eq:classifier:1}
F: \ \mathcal{U}\rightarrow \mathcal{L},
\end{equation}
mapping elements of an input set $\mathcal{U}$ into a finite label set $\mathcal{L}$
\begin{equation}\label{eq:classifier:2}
\mathcal{L}=\{\ell_1,\ell_2,\dots,\ell_q\}, \ q\in\Natural.
\end{equation}
We assume that the map $F$ is measurable,
and that we are given a measurable map $\Phi$
\begin{equation}\label{eq:feature_map}
\Phi: \ \mathcal{U}\rightarrow \Real^d, \ d\in\Natural,
\end{equation}
assigning an element of $\Real^d$ to an element from $\mathcal{U}$. The map $\Phi$ can be thought of as a feature or an observation map allowing elements $u$ of $\mathcal{U}$ to be represented by $\Phi(u)$ from $\Real^d$. 

As an example, consider a standard image classification problem. In this case, the set $\mathcal{U}$ is the set of all images of fixed size and colour depth, and the set $\mathcal{L}$ is the set of labels corresponding to relevant image classes, and $F$ can be any appropriately designed classifier, including deep convolutional neural networks. The map $\Phi$ can be any representation of an image by an element from $\Real^d$, for example through the outputs of the neural network's hidden layers.

We also assume that we have access to data that can be used to construct AI error correctors, in the form of a finite multi-set
\begin{equation}\label{eq:training_set}
\mathcal{S} = \{ (u_i,y_i) \ | \ u_i\in\mathcal{U}, \ y_i\in\mathcal{L}, \ i=1,\dots,M\},
\end{equation}
where $M\in\Natural$ is the cardinality of $\mathcal{S}$. Elements of $\mathcal{S}$ are sampled independently from some {\it unknown} probability distribution $P_{D}$ defined on $\mathcal{U}\times\mathcal{L}$. We assume that $\mathcal{S}$ has not been used for the choice of the classifier $F$ or the observation/feature map $\Phi$. Moreover, the sets
\[
\mathcal{S}_j=\{(u,y)\in\mathcal{S} \,| \, F(u)=\ell_j\}
\]
are not empty for all $j=1,\dots,q=|\mathcal{L}|$. Finally, let
\[
\reject: \ \reject\notin\mathcal{L}
\]
be a label that does not belong to the set $\mathcal{L}$, which we use to denote an AI correctors' output corresponding to rejecting the decision of the classifier $F$.

In this paper (Section~\ref{sec:Main}), we present a solution to the following problem. 

\begin{prob}[Correction of AI errors]\label{prob:error_corrector} 
Suppose we are given an AI system $F$, feature map $\Phi$, and corrector training set $\mathcal{S}$ as described above.
Find an error correcting map $\mathcal{A}:\mathcal{L}\times\Real^d\rightarrow\mathcal{L}\cup\{\reject\}$ and positive numbers $\upsilon_j\in[0,1]$, $\gamma_j\in[0,1]$, $j=1,\dots,|\mathcal{L}|$, such that for any $(u,\ell)\sim P_D$
\begin{enumerate}
\item the corrector $\mathcal{A}$ accepts a correct decision of the underlying AI system with probability at least $\upsilon_j$, i.e.
\[
P(\mathcal{A}(F(u),\Phi(u))=\ell_j \,| \, F(u)=\ell_j, \ \ell=\ell_j ) \geq \upsilon_j
\]
\item the corrector $\mathcal{A}$ rejects an incorrect decision of the underlying AI system with probability at least $\gamma_j$, i.e.
\[
P(\mathcal{A}(F(u),\Phi(u))=\reject \,| \, F(u)= \ell_j, \ \ell\neq \ell_j ) \geq \gamma_j.
\]
\end{enumerate}
\end{prob}

The first group of events
\[
\mathcal{A}(F(u),\Phi(u))=\ell_j,  \ F(u)=\ell_j, \ \ell=\ell_j
\]
corresponds to the case when the AI correcting map $\mathcal{A}$ returns the same label as the underlying classifier $F$ provided that the classifier correctly produces the label $\ell_j$.
Here $\upsilon_j$ measures the extent to which the corrector does not interfere with correct decisions of the original classifier $F$ (larger values are better).

The second group of events
\[
\mathcal{A}(F(u),\Phi(u))=\reject, \ F(u)= \ell_j, \ \ell\neq \ell_j 
\]
corresponds to the case when the corrector successfully detects and reports an error made by the underlying classifier $F$.
The value of $\gamma_j$ therefore measures the corrector's sensitivity to errors made by the underlying system (larger values are better). 

Problem~\ref{prob:error_corrector} also expresses our choice to work with separate probability bounds for each class label $\ell_j$, due to the potential utility of this added granularity in applications.
If information about probabilities of each label $\ell_j$ occurring is known, the label-specific bounds in Problem \ref{prob:error_corrector} can be collapsed into just two overall bounds on interference and successful identification of errors, regardless of the algorithm used:
\[
\begin{split}
&P(\mathcal{A}(F(u),\Phi(u))=\ell |  \ F(u)=\ell) =\\
& \sum_{j=1}^{q}P(\mathcal{A}(F(u),\Phi(u))=\ell_j |  \ F(u)=\ell_j, \ \ell=\ell_j) P(\ell=\ell_j)\\
&   \geq \sum_{j=1}^q P(\ell=\ell_j)\upsilon_j,
\end{split}
\] 
and
\[
\begin{split}
&P(\mathcal{A}(F(u),\Phi(u))=\reject | \ F(u)\neq \ell) =\\
&\sum_{j=1}^{q}P(\mathcal{A}(F(u),\Phi(u))=\reject |  \ F(u)=\ell_j, \ \ell\neq \ell_j) P(\ell\neq \ell_j)\\
& \geq  \sum_{j=1}^q (1-P(\ell=\ell_j))\gamma_j.
\end{split}
\]
In this sense, Problem~\ref{prob:error_corrector} is more general and detailed than a similar problem requiring guarantees on just these two (class-independent) probabilities.

\section{Main Results}\label{sec:Main}

In this section, we present an algorithm for solving Problem~\ref{prob:error_corrector} and bounds on the probability of success, built around low dimensional projections of representations of the data.
Algorithm~\ref{alg:corrector_general} summarises the proposed method, and Theorem~\ref{thm:main_corrector_bounds} presents corresponding bounds on the probability that the corrector correctly accepts or rejects the underlying system's decisions.

The idea of exploring low-dimensional projections for correcting AI errors is consistent with previously proposed approaches \cite{gorban2021high,gorban2018blessing,gorban2018correction,gorban2019one} and \cite{kuleshov2022calibrated}. 
The works \cite{gorban2021high,gorban2018blessing,gorban2018correction,gorban2019one} only considered linear projections of data representations living in essentially high- (and possibly infinite-) dimensional feature spaces. Here we consider the problem in greater generality and require neither high dimensionality or the linearity of the projection maps. In contrast to \cite{kuleshov2022calibrated}, where a related problem of regression calibration is considered, we aim to provide rigorous non-asymptotic and \emph{a priori} computable probability bounds on the decisions of the corrector. 

\subsection{Theory}

To fix notation, let $\mathcal{H}$ denote a (finite or infinite) class of measurable functions from $\Real^d$ to $\Real$, representing projection maps for elements $\Phi(u)$, $u\in\mathcal{U}$ onto low-dimensional sub-spaces. 
We split the set $\mathcal{S}$ of corrector training examples into those examples from each class which have been correctly or incorrectly classified by the underlying AI system, defined respectively as
\[
\begin{split}
\mathcal{S}_{+,j}=&\{(u,y)\in\mathcal{S}_j \ | \ y=\ell_j\},\\
\mathcal{S}_{-,j}=&\{(u,y)\in\mathcal{S}_j \ | \ y\neq \ell_j\},
\end{split}
\]
with cardinalities $M_{+,j} = |\mathcal{S}_{+,j}|$ and $M_{-,j} = |\mathcal{S}_{-,j}|$.
From these, and a given projector $h_j \in \mathcal{H}$ for each class $j$, we construct 
the empirical cumulative distribution functions
\begin{equation}\label{eq:empirical_cumulative}
\begin{split}
F_{+,j}(s) = &\frac{1}{M_{+,j}}\sum_{(u, y) \in \mathcal{S}_{+,j}} \boldsymbol{1}_{h_j(\Phi(u)) \leq s}, \\
F_{-,j}(s) = &\frac{1}{M_{-,j}}\sum_{(u, y) \in \mathcal{S}_{-,j}} \boldsymbol{1}_{h_j(\Phi(u)) \leq s}.
\end{split}
\end{equation}

\begin{algorithm}\caption{AI Error Corrector}\label{alg:corrector_general}
\small
\SetKwData{Left}{left}\SetKwData{This}{this}\SetKwData{Up}{up}
\SetKwFunction{Union}{Union}\SetKwFunction{FindCompress}{FindCompress}
\SetKwInOut{Input}{Input}\SetKwInOut{Output}{Output}

\Input{Classifier $F$, feature map $\Phi$, training set $\mathcal{S}$, projectors $\{h_j \in \mathcal{H}, \ j=1,\dots,q\}$ and a positive real numbers $\Delta_j\in(0,1)$, $j=1,\dots,q$.}
\BlankLine
\For{$j\leftarrow 1$ \KwTo $q$}{
Build the subset $\mathcal{S}_{-,j}$ of $\mathcal{S}$




Define $\mathcal{A}_j: \Real^d \rightarrow \mathcal{L}\cup\{\reject\}$ as\label{alg:line_A_j}
\[
\mathcal{A}_j (z)=
\begin{cases}
\reject &\text{if } h_j(z) \leq F_{-,j}^{\dagger}(\Delta_j), \\
\ell_j &\text{otherwise}.
\end{cases}
\]
}

Define\label{alg:line_A}
$\mathcal{A}: \mathcal{L}\times\Real^d\rightarrow  \mathcal{L}\cup\{\reject\}$ for each $\ell_j \in \mathcal{L}$ as
\[
\mathcal{A}(\ell_j, z)= A_j(z).
\]

\Output{AI corrector map $\mathcal{A}$. }
\end{algorithm}

\begin{thm}\label{thm:main_corrector_bounds} Let the AI system $F$, feature map $\Phi$, and corrector training set $\mathcal{S}$ be defined as in~\eqref{eq:classifier:1}--\eqref{eq:training_set}. Suppose that the elements of $\mathcal{S}$ are independently sampled from the (unknown) data distribution $P_D$, that $\mathcal{S}_{+,j}$ and $\mathcal{S}_{-,j}$ are not empty, and that $F$ and $\Phi$ are chosen independently of $\mathcal{S}$.

For each class index $j$, select $h_j$ from $\mathcal{H}$ independently of $\mathcal{S}$, and let $\mathcal{A}$ be the output of Algorithm~\ref{alg:corrector_general} with inputs $F,\Phi,\mathcal{S}, \{h_j\in\mathcal{H}| j=1,\dots,q\}$ and $\Delta_j\in(0,1)$.
Then, for a new independent sample $(u, \ell)\sim P_D$ and each $j=1,\dots,q$,
\[
\begin{split}
& P(\mathcal{A}(F(u),\Phi(u))=\ell_j\,| \, F(u)=\ell_j, \ \ell=\ell_j ) \geq\\
& \ \ \ \ \ \ \ \ \ \ \ \ \ \ \ \ \ \ \ \ \ \ \ \ \ \ \ 1 - \psi(F_{+,j}(F^{\dagger}_{-,j}(\Delta_j)), M_{+,j}),
\end{split}
\]
and
\[
\begin{split}
& P(\mathcal{A}(F(u),\Phi(u))=\reject \,| \, F(u)= \ell_j, \ \ell\neq \ell_j ) \geq
\rho(\Delta_j,M_{-,j}),
\end{split}
\]
where $\rho,\psi:[0,1]\times\Natural\rightarrow \Real$ are defined as
\[
\rho(a,d)=\sup_{\varepsilon\in(0,1]} \max\{a - \varepsilon,0\}(1 - 2 \exp(-2 d \varepsilon^2)),
\]
\[
\psi(a,d)=\inf_{\varepsilon\in(0,1]}  2\exp(-2d\varepsilon^2)+\min\{1,a+\varepsilon\}.
\]
\end{thm}
{\it Proof of Theorem \ref{thm:main_corrector_bounds}.} Given that $F$ is measurable and elements in $\mathcal{S}$ are independently sampled from the distribution $P_D$, we can conclude that $\mathcal{S}_j$ and $\mathcal{S}_{+,j}$, $\mathcal{S}_{-,j}$ are all independent samples from the conditional distributions $P_D((u,y) | F(u)=\ell_j)$, $P_D((u,y) | y=\ell_j, \ F(u)=\ell_j)$, and $P((u,y)|y\neq \ell_j, \ F(u)=\ell_j)$. 
Moreover, since $h_j\in\mathcal{H}$ is chosen independently of $\mathcal{S}$ and that $h_j$ and $\Phi$ are measurable, we may conclude that elements of 
the sets
\[
\begin{split}
\mathcal{X}_{+,j}=&\{h_j(\Phi(u)), \text{ for each } (u,y) \in \mathcal{S}_{+,j}\},\\
\mathcal{X}_{-,j}=&\{h_j(\Phi(u)), \text{ for each } (u,y) \in \mathcal{S}_{-,j}\}.\\
\end{split}
\]
are independent samples from the respective conditional probability distributions induced on $\Real$ by $h_j$.

To proceed with the proof we will need the following lemma.

\begin{lem}\label{lem:empirical_probability} Let $\theta\in\Real$ and $\mathcal{X}\subset\mathcal{\Real}$ be a finite multi-set whose elements are sampled independently from some unknown probability distribution. Let $|\mathcal{X}|=n$, and $F_n$ be defined as follows:
\[
F_n(s)=\frac{1}{n}\sum_{x\in\mathcal{X}}\boldsymbol{1}_{x\leq s}.
\]
Let $z$ be a new independent sample from the same distribution and consider the events:
\[
E_1: \ z\leq \theta, \ E_2: \ z> \theta.
\]

Then
\begin{align}
\psi(F_n(\theta),n) \geq  & P(E_1) \geq \rho(F_n(\theta),n) 
\label{eq:probability_bounds:E1}
\\
1 - \rho(F_n(\theta),n)  \geq  & P(E_2) \geq  1 - \psi(F_n(\theta),n).\label{eq:probability_bounds:E2}
\end{align}
\end{lem}
{\it Proof of Lemma \ref{lem:empirical_probability}.} Consider the event $E_1$, and let $P_c$ be the cumulative distribution function associated with the unknown probability distribution from which both the multi-set $\mathcal{X}$ and $z$ were independently drawn. It is clear that 
\[
P(E_1)=P_c(\theta), \ P(E_2)=1-P_c(\theta).
\]
Introducing the auxiliary event
\[
E(\varepsilon): \ \sup_{s\in\Real}|F_n(s)-P_c(s)|>\varepsilon,
\]
and its complement $\bar{E}(\varepsilon): \operatorname{not} E(\varepsilon)$.
The Dvoretzky–Kiefer–Wolfowitz inequality~\cite{DKW,Massart} states that, for any $\varepsilon > 0$,
\begin{equation}\label{eq:KDW}
P\left(\sup_{s\in\Real}|F_n(s)-P_c(s)|>\varepsilon \right) \leq 2 \exp(-2 n \varepsilon^2),
\end{equation}
and therefore 
$
P(E(\varepsilon)) \leq 2 \exp(-2 n \varepsilon^2) 
$
and
$
P(\bar{E}(\varepsilon))
\geq 1 - 2 \exp(-2 n \varepsilon^2)
$.

By the law of total probability, we may express
\begin{equation}\label{eq:total_probability}
P(E_1)=P(E_1 \, | \, E(\varepsilon)) P (E(\varepsilon)) + P(E_1 \, | \, \bar{E}(\varepsilon)) P (\bar{E}(\varepsilon)),
\end{equation}
and hence
\begin{equation}\label{eq:bound_lower}
P (E_1) \geq  P(E_1 \, | \, \bar{E}(\varepsilon)) P(\bar{E}(\varepsilon)).
\end{equation}
The fact that
\begin{equation}\label{eq:conditional_bounds}
\begin{split}
\min\{1,F_n(\theta) + \varepsilon\}  &\geq P(E_1 \, | \, \bar{E}(\varepsilon)) 
\geq \max\{F_n(\theta) - \varepsilon,0\},
\end{split}
\end{equation}
therefore implies 
\[
P (E_1) \geq \max\{F_n(\theta) - \varepsilon,0\}(1 - 2 \exp(-2 n \varepsilon^2)).
\]
Similarly, \eqref{eq:total_probability} also implies that
\[
P(E_1)\leq P(E(\varepsilon))+P(E_1|\bar{E}(\varepsilon)),
\]
and it follows by applying~\eqref{eq:KDW} and \eqref{eq:conditional_bounds} that
\[
P(E_1) \leq 2\exp(-2n\varepsilon^2)+\min\{1,F_n(\theta)+\varepsilon\}.
\]
Taking the supremum with respect to $\varepsilon$ for the lower bound and the infimum for the upper bound produces the desired bounds on the probability of event $E_1$.

Given that
\[
P(E_2)=1-P(E_1),
\]
bounds on $P(E_2)$ follow from the bounds on $P(E_1)$ $\square$.

\textit{Proof of Theorem~\ref{thm:main_corrector_bounds} continued.}
By the construction of $\mathcal{A}$ in Algorithm~\ref{alg:corrector_general} (lines \ref{alg:line_A_j}, \ref{alg:line_A}), the probability of the corrector correctly accepting the label assigned by the underlying AI system is given by
\begin{align}
&P(\mathcal{A}(F(u),\Phi(u))=\ell_j \,| \, F(u)=\ell_j, \ \ell=\ell_j)
\notag
\\&\quad
=
P(h_j(\Phi(u))>F^{\dagger}_{-,j}(\Delta_j)\,| \, F(u)=\ell_j, \ y=\ell_j).
\label{eq:proxy_probability_true_positive}
\end{align}
Since $F_{+,j}$ is defined in~\eqref{eq:empirical_cumulative} to be the empirical approximation of the cumulative distribution function of the conditional distribution $P(h_j(\Phi(u))\,| \, F(u)=\ell_j, \  y=\ell_j)$, bound~\eqref{eq:probability_bounds:E2} from Lemma \ref{lem:empirical_probability} (with $\theta = F^{\dagger}_{-,j}(\Delta_j)$) implies that
\begin{equation}\label{eq:proof_bound_1}
\begin{split}
 & P(h_j(\Phi(u))>F^{\dagger}_{-,j}(\Delta_j)\,| \, F(u)=\ell_j, \ y=\ell_j)  \geq \\
& \ \ \ \ \ \ \ \ \ 1 - \psi(F_{+,j}(F^{\dagger}_{-,j}(\Delta_j), M_{+,j}).
\end{split}
\end{equation}

Turning to the probability of the corrector correctly rejecting an incorrect label assigned by the underlying AI system, we observe that
\begin{align*}
&P(\mathcal{A}(F(u),\Phi(u))=\reject \,| \, F(u)=\ell_j, \ \ell\neq\ell_j)
\\&\quad
=
P(h_j(\Phi(u))\leq F^{\dagger}_{-,j}(\Delta_j) \,| \, F(u)=\ell_j, \ y\neq\ell_j).
\end{align*}
Similarly to the previous case, we note that the conditional distribution  $P(h(\Phi(u))| \ F(u)=\ell_j, \ y\neq\ell_j)$ admits a cumulative distribution function with empirical approximation $F_{-,j}$ which is used by Algorithm \ref{alg:corrector_general}.
Invoking bound~\eqref{eq:probability_bounds:E1} of Lemma \ref{lem:empirical_probability} therefore establishes the bound
\begin{equation}\label{eq:proof_bound_2}
\begin{split}
&P(h_j(\Phi(u))\leq F^{\dagger}_{-,j}(\Delta_j) \,| \, F(u)=\ell_j, \ y\neq\ell_j)\geq\\ 
&\ \ \ \  \rho(F_{-,j}(F^{\dagger}_{-,j}(\Delta_j)),M_{-,j})=\rho(\Delta_j,M_{-,j}),
\end{split}
\end{equation}
and combining (\ref{eq:proof_bound_1}) and (\ref{eq:proof_bound_2}) completes the proof. $\square$

\subsection{Discussion}
Theorem \ref{thm:main_corrector_bounds} provides estimates of the probability that the AI error corrector correctly accepts or rejects the responses of the underlying AI system.
The bounds have some particular features, which we discuss here.

\subsubsection{Computing the bounds}
The bounds depend on the sizes $M_{-,j}$, $M_{+,j}$ of the corrector's training datasets $\mathcal{S}_{-,j}$, $\mathcal{S}_{+,j}$ and the threshold parameters $\Delta_j$ used to construct the corrector.
Although the bound on the probability of accepting correct predictions made by the underlying AI system depends on the empirical cumulative distribution function $F_{+,j}$, the bound on the probability of correctly rejecting an incorrect decision can be computed \emph{a priori} just from $M_{-,j}$ and the design parameters $\Delta_j$, without the knowledge of either $F_{+,j}$ or $F_{-,j}$.
Given a fixed-size set of training data, setting an acceptable guarantee on the probability of incorrectly rejecting a sample (using Theorem~\ref{thm:main_corrector_bounds}) could therefore be used as a design principle for selecting the parameters $\Delta_j$.

\subsubsection{Tightness of the bounds}
\begin{figure}
\includegraphics[width=\linewidth]{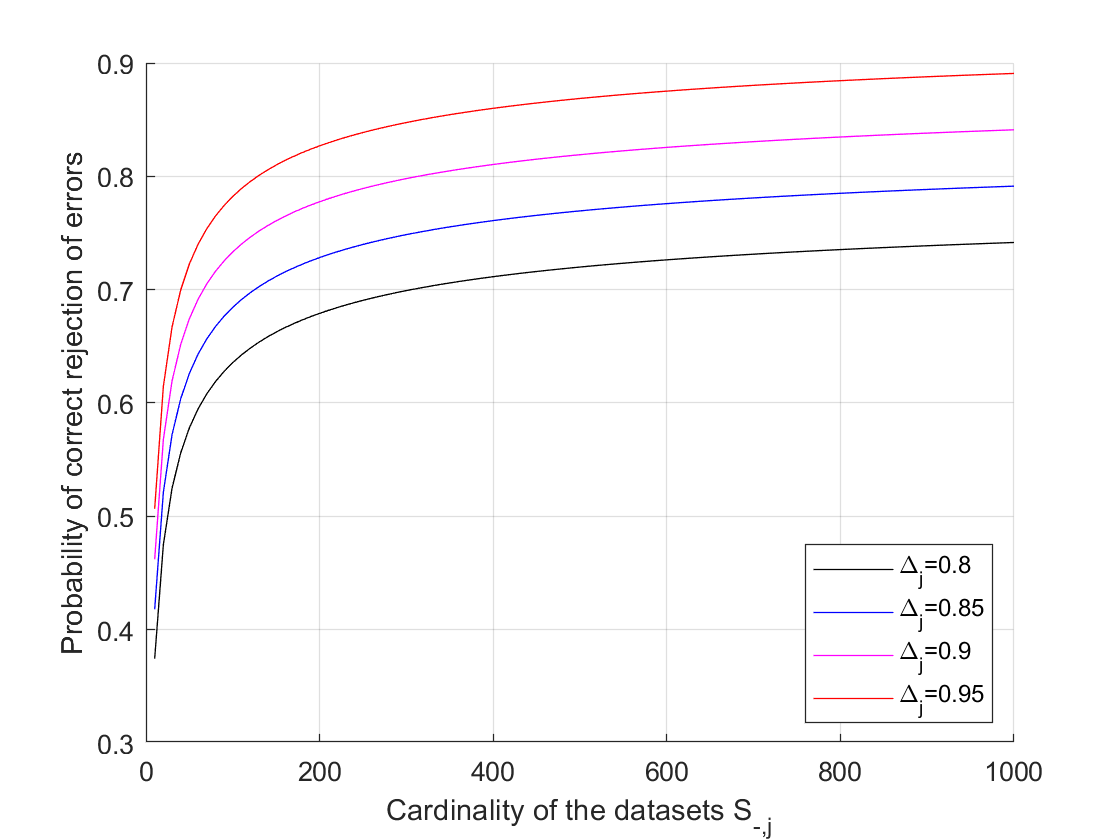}
\caption{Lower bound on the probability of correct rejection of errors, $P (\mathcal{A}(F (u), \Phi(u)) = \reject| F (u) = \ell_j , \ \ell\neq\ell_j)$, provided in Theorem \ref{thm:main_corrector_bounds}  and expressed as a function of the cardinality $M_{-,j}$ of the set $\mathcal{S}_{-,j}$ for different values of $\Delta_j=0.8,0.85,0.9,0.95$.}\label{fig:rho_bound}
\end{figure}

The bound on the correct rejection probability is plotted in Figure~\ref{fig:rho_bound} as functions of $M_{-,j}$ for different values of $\Delta_j$.
The relatively slow convergence of the bounds with $M_{-,j}$ to their limiting values is due to the reliance on the Dvoretzky–Kiefer–Wolfowitz-Masser inequality \eqref{eq:KDW}.
This result produces distribution agnostic bounds, meaning that they do not depend on any assumptions on the data distribution, but this flexibility is comes at the price of \emph{worst-case} bounds.
The bound~\eqref{eq:KDW} is known to be tight, however, and can only be improved if extra information about the underlying distribution is available. 

\subsubsection{Abstaining from making a decision}
An immediate consequence of Theorem \ref{thm:main_corrector_bounds} and a potential application of Algorithm \ref{alg:corrector_general} is the possibility to use AI error correctors as a basis for abstaining from making classification decisions.
Indeed, by appropriately choosing the values of $\Delta_j$ one can ensure that only answers with certain confidence levels are accepted.
Samples of input data not passing this threshold could then be forwarded to a human expert or to an alternative AI model.

\section{Example: AI correctors for pottery classification}\label{sec:Example}

In this section, we illustrate the application of the approach to the problem of pottery classification from images of broken pieces of ceramic material.

The problem of image classification has been a key driver of the AI revolution over the last decade.
Since the celebrated success of convolutional neural networks on ImageNet \cite{krizhevsky2012imagenet}, these and other AI models have been shown to be successful in numerous other applications and problems such as autonomous driving \cite{gupta2021deep} and cancer diagnosis \cite{chen2023development}.
Despite these successes, there are classes of problems where applying state-of-the-art AI tools such as deep neural networks remains a challenge. 

One such challenging problem is the task of identification of archaeological artefacts from images of archaeological remains in the form of broken ceramic pieces known as \emph{sherds}.
The difficulty of the problem is emphasised by the fact that even the identification of complete or near-complete vessels from photographs is a challenging problem~\cite{nunez2021learning}, even though these images contain significantly more information about the object than images of sherds do.
Compounded with this is the difficulty of obtaining training data, since the volume of labelled images of sherds of any particular class of artefact is limited.
The question we have explored here is how the AI correctors developed in Section~\ref{sec:Main} could be used to help with deriving a suitable solution.

\subsection{Core AI classifier}

As a core AI classifier, we used InceptionV3 deep neural network initialised by the ImageNet weights and the following structure of the additional layers we introduced after removing of the classification layer in the original net: Dropout(0.5), GlobalAveragePooling2D, Dense(512, activation=’relu’), Dense(5, activation=’softmax)’.

\subsection{Data}

\subsubsection{Data for training core classifier}

The data we had available for building our core deep learning classifier comrised images of ceramic pieces from Dragendorff \cite{webster1996roman} forms Dr27 (class 1), Dr33 (class 2), Dr35 (class 3),  Dr37 (class 4), and Dr38 (class 5). The composition of the images of real sherds is provided in Table \ref{tab:Training_data_real}. Examples of images of real sherds are shown in Fig. 

\begin{figure*}
\centering
\includegraphics[width=0.15\textwidth]{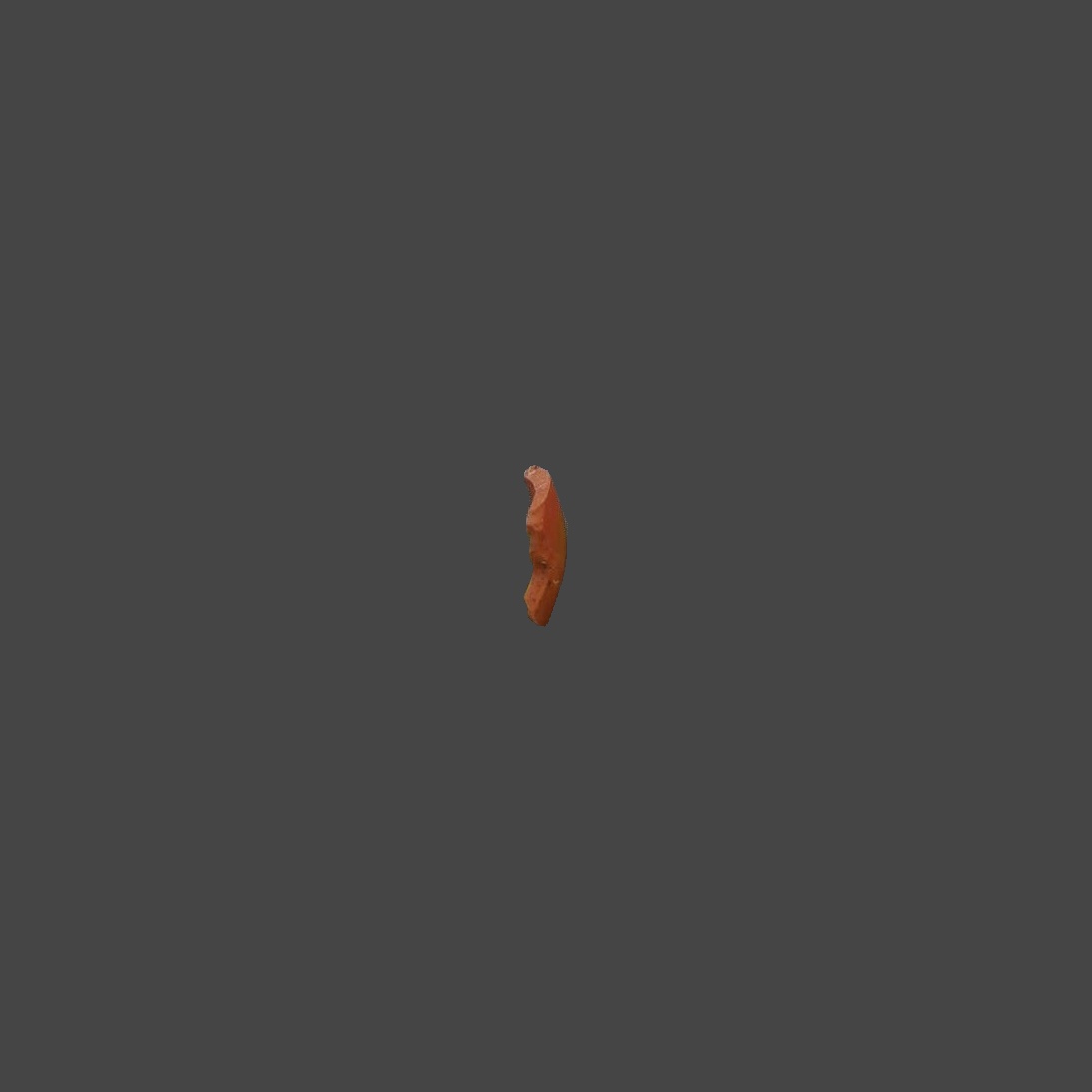}\hspace{1mm}\includegraphics[width=0.15\textwidth]{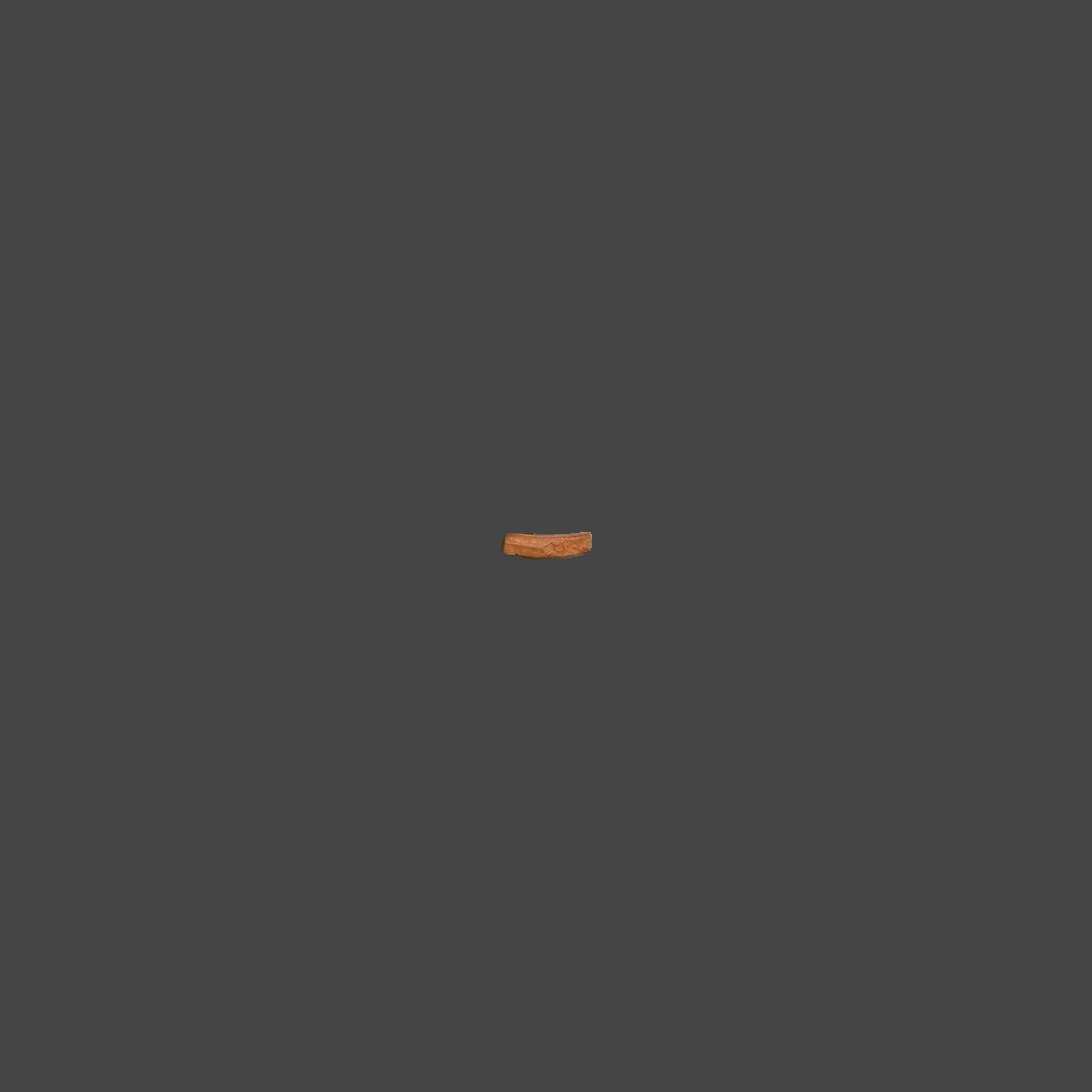}\hspace{1mm}\includegraphics[width=0.15\textwidth]{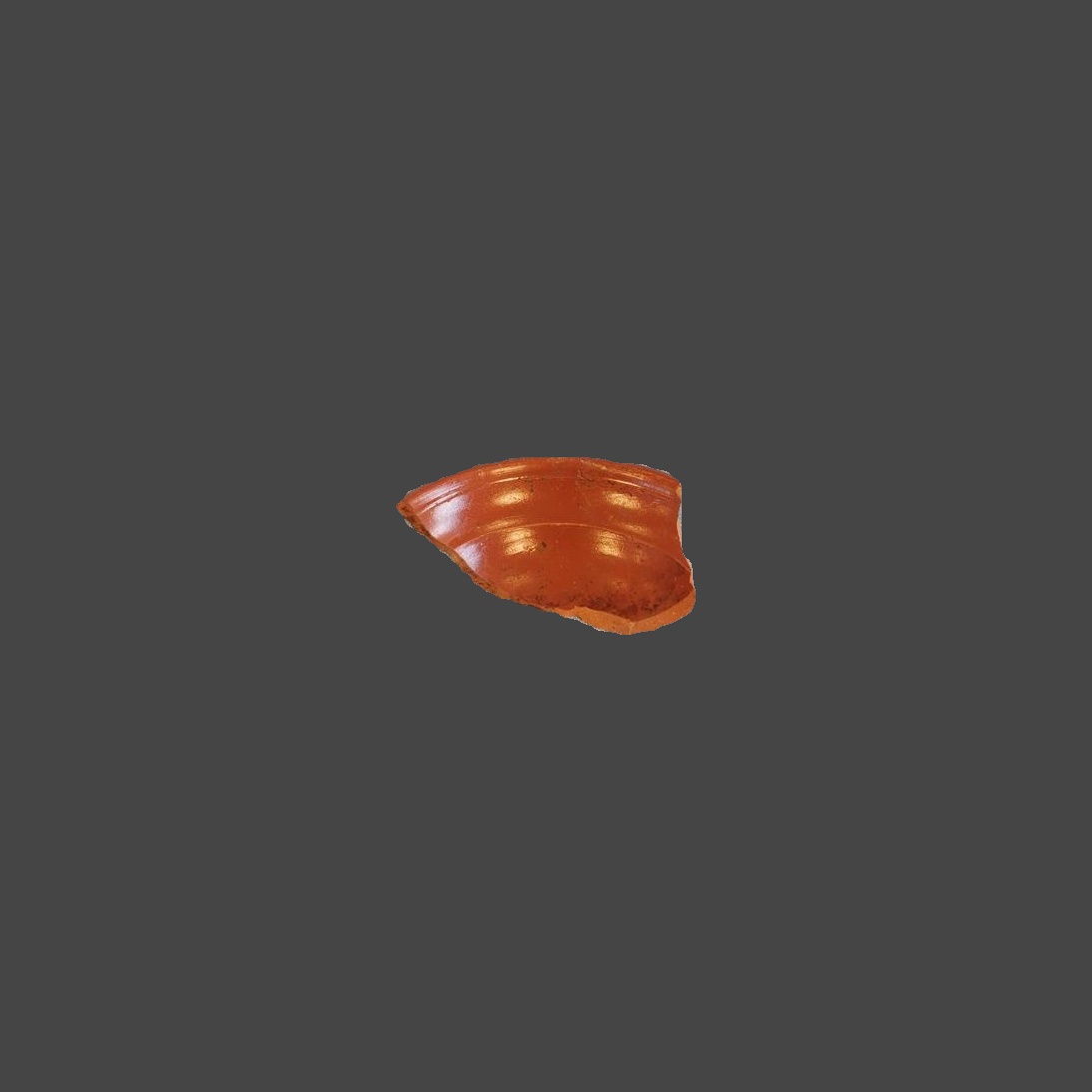}\hspace{1mm}\includegraphics[width=0.15\textwidth]{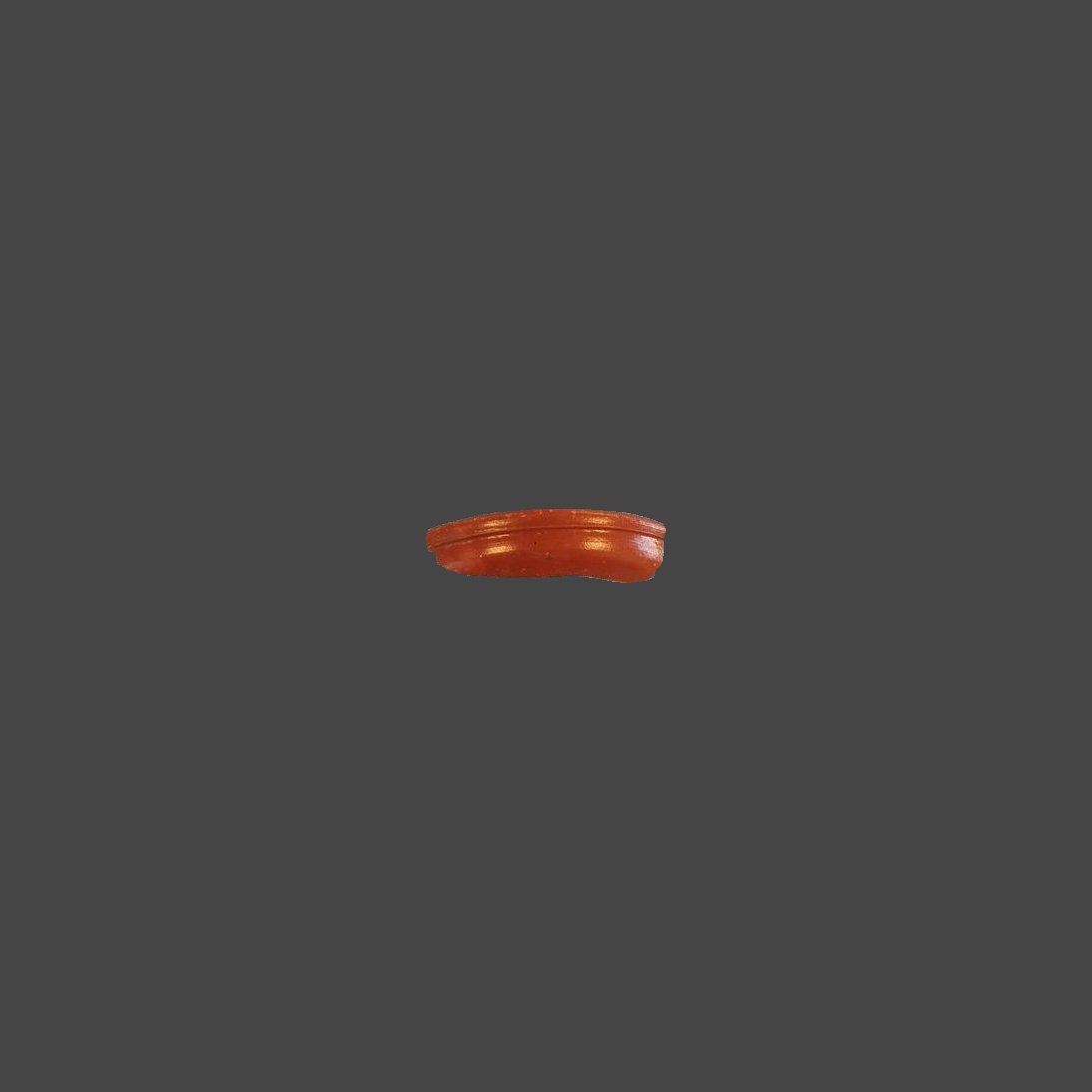}\hspace{1mm}\includegraphics[width=0.15\textwidth]{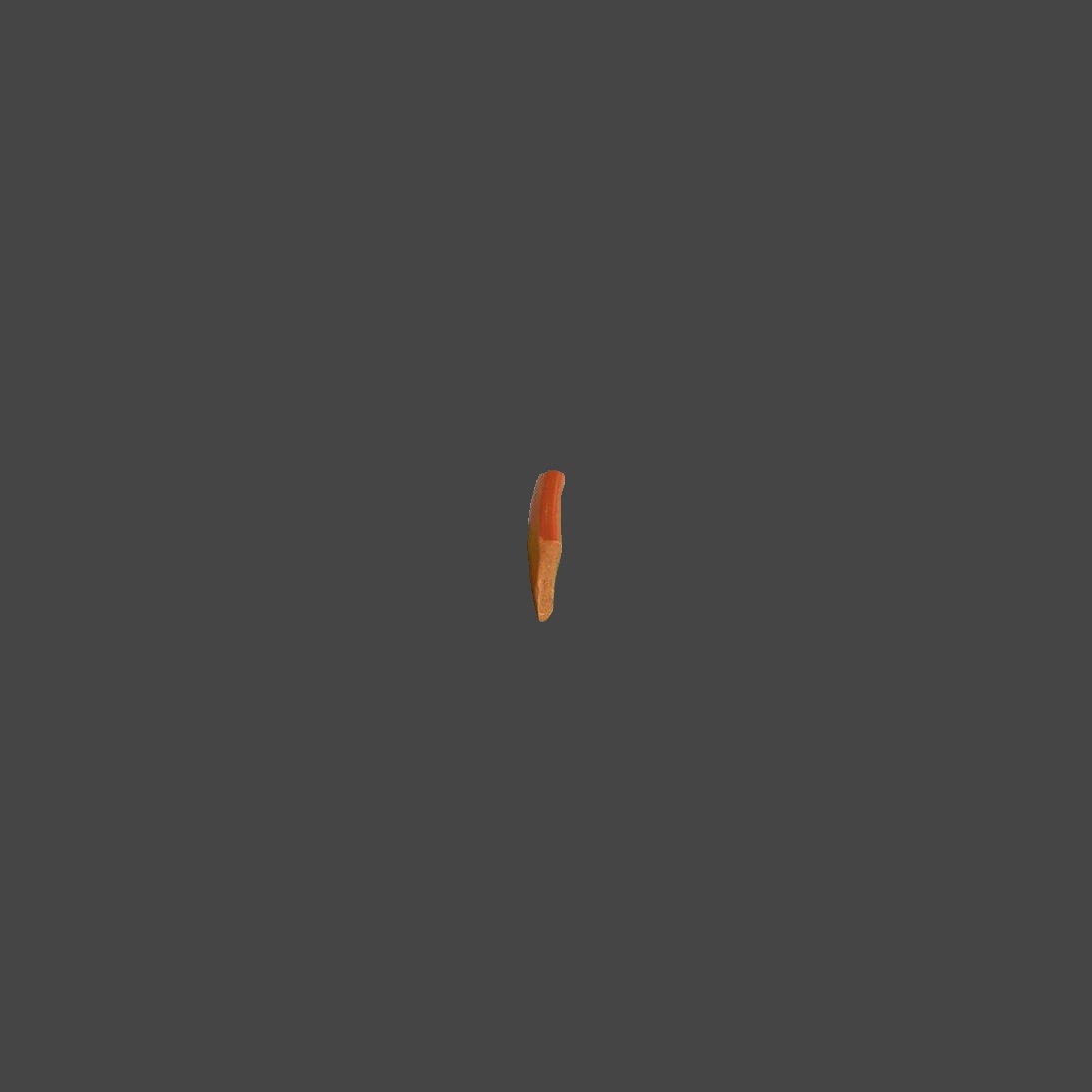}\\
\vspace{2mm}
\includegraphics[width=0.15\textwidth]{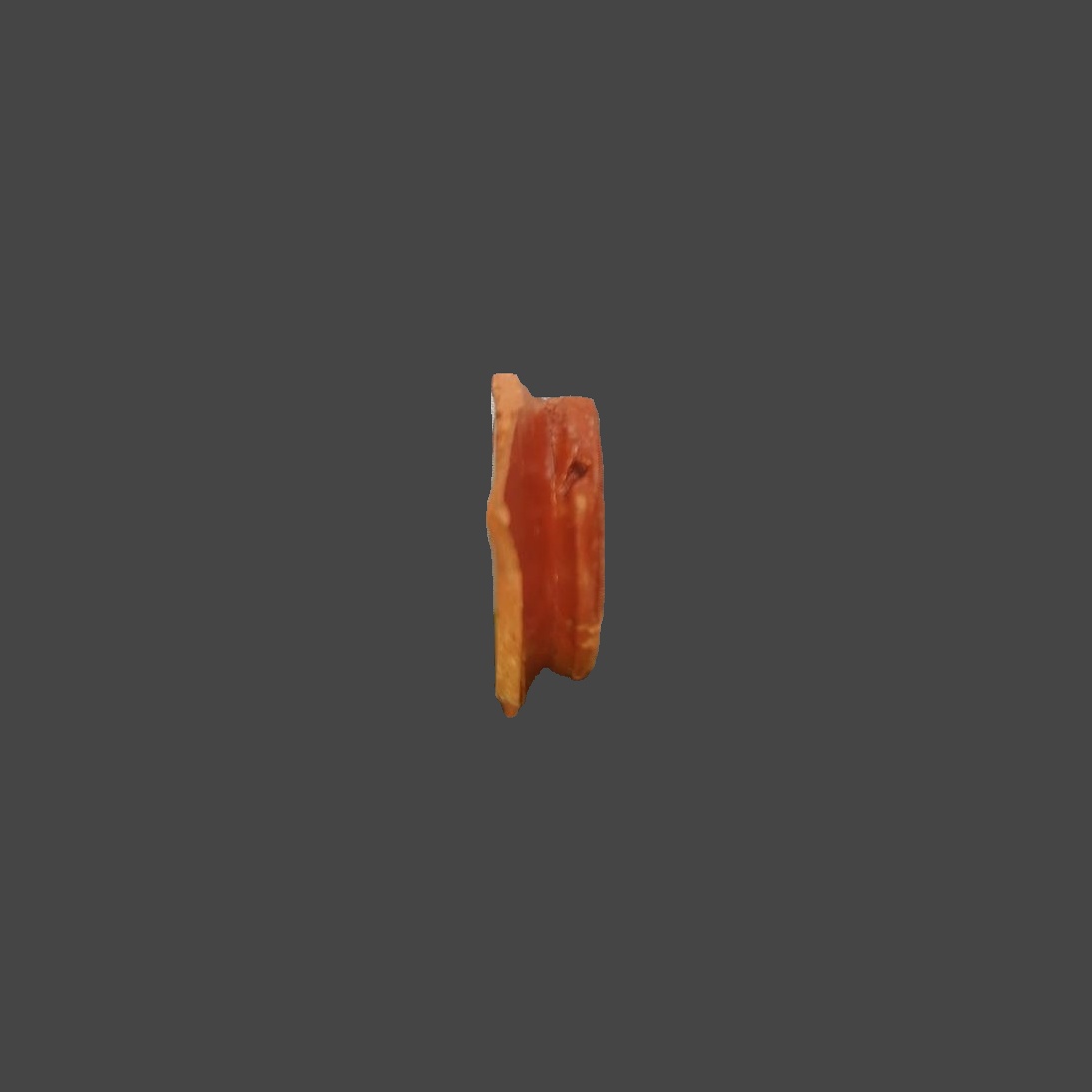}\hspace{1mm}\includegraphics[width=0.15\textwidth]{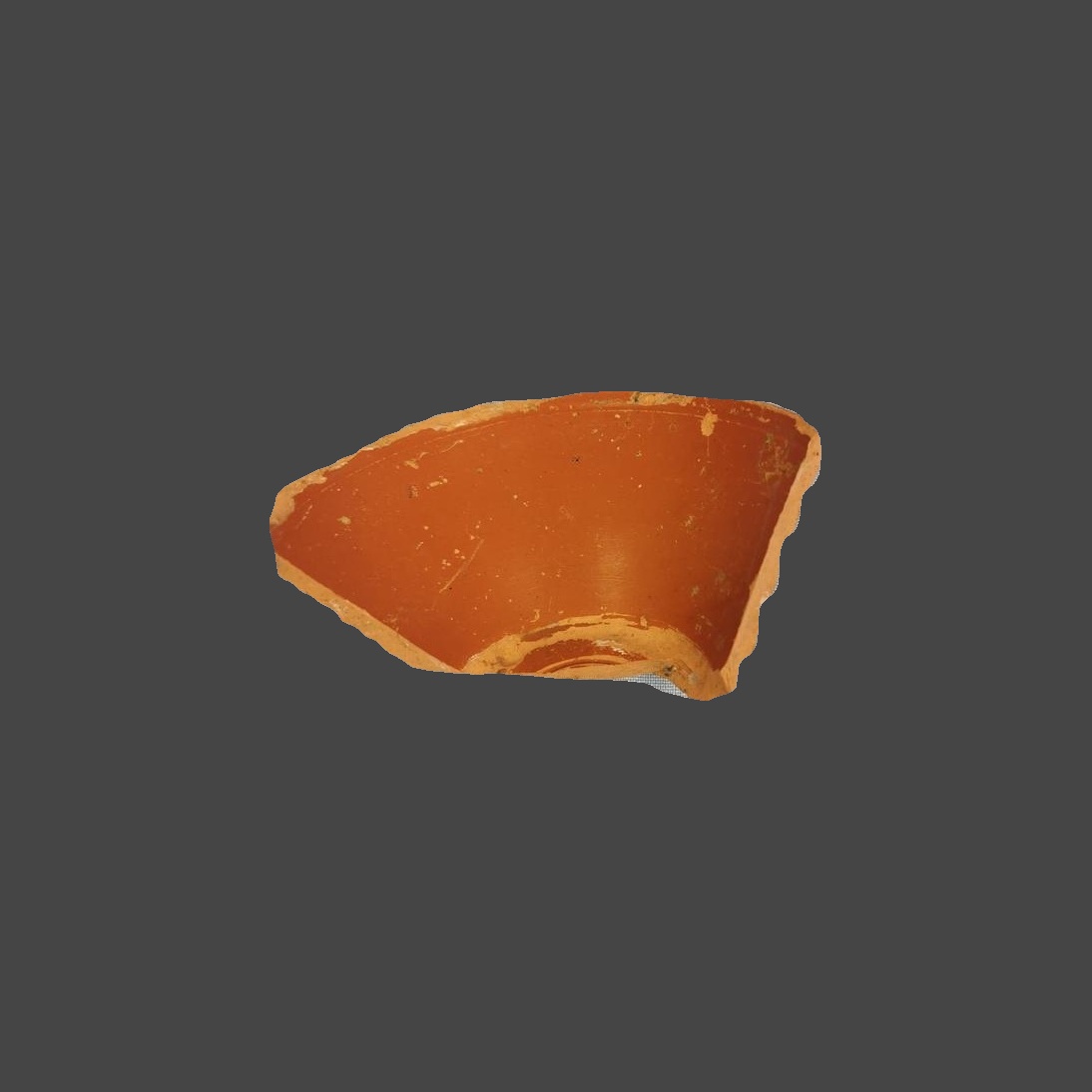}\hspace{1mm}\includegraphics[width=0.15\textwidth]{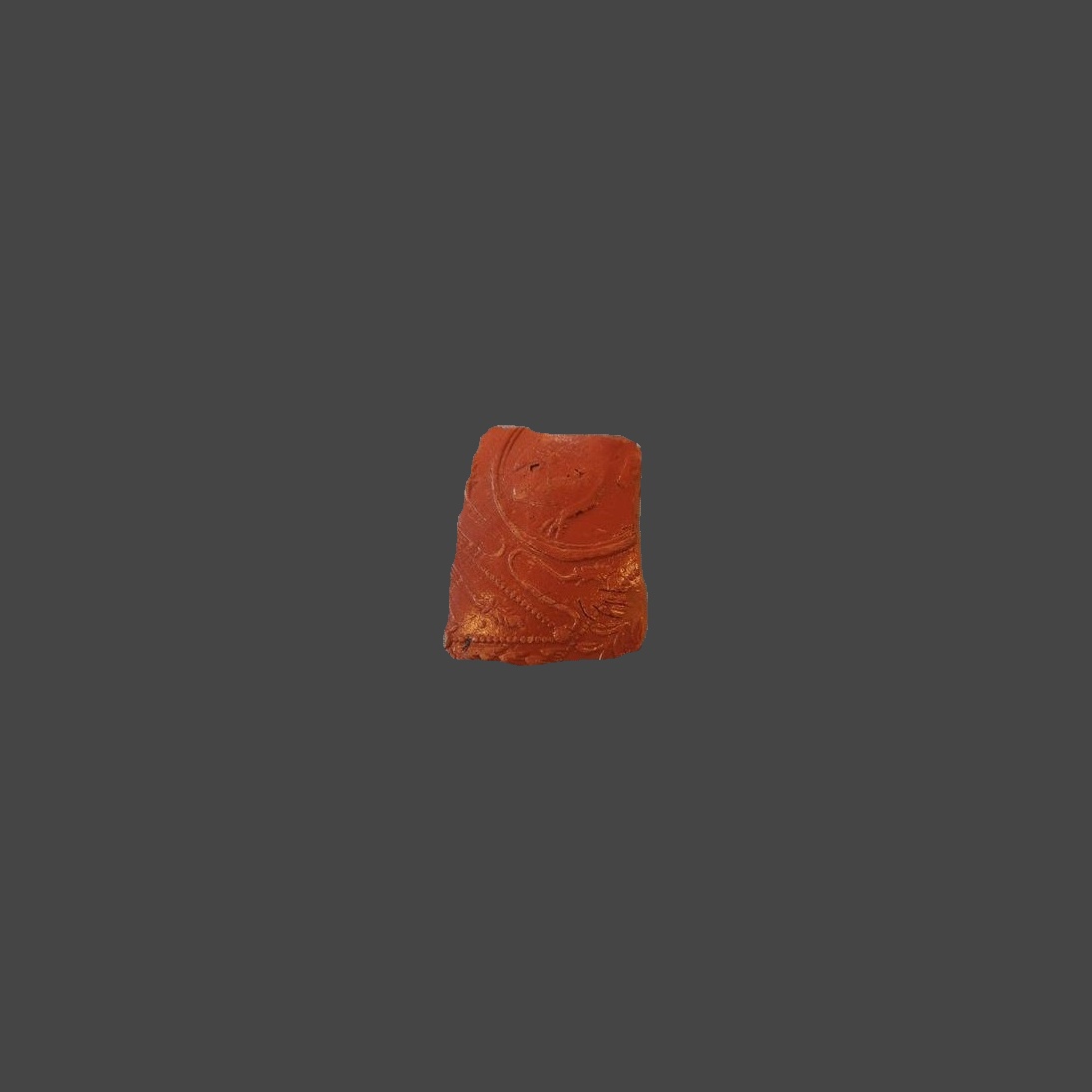}\hspace{1mm}\includegraphics[width=0.15\textwidth]{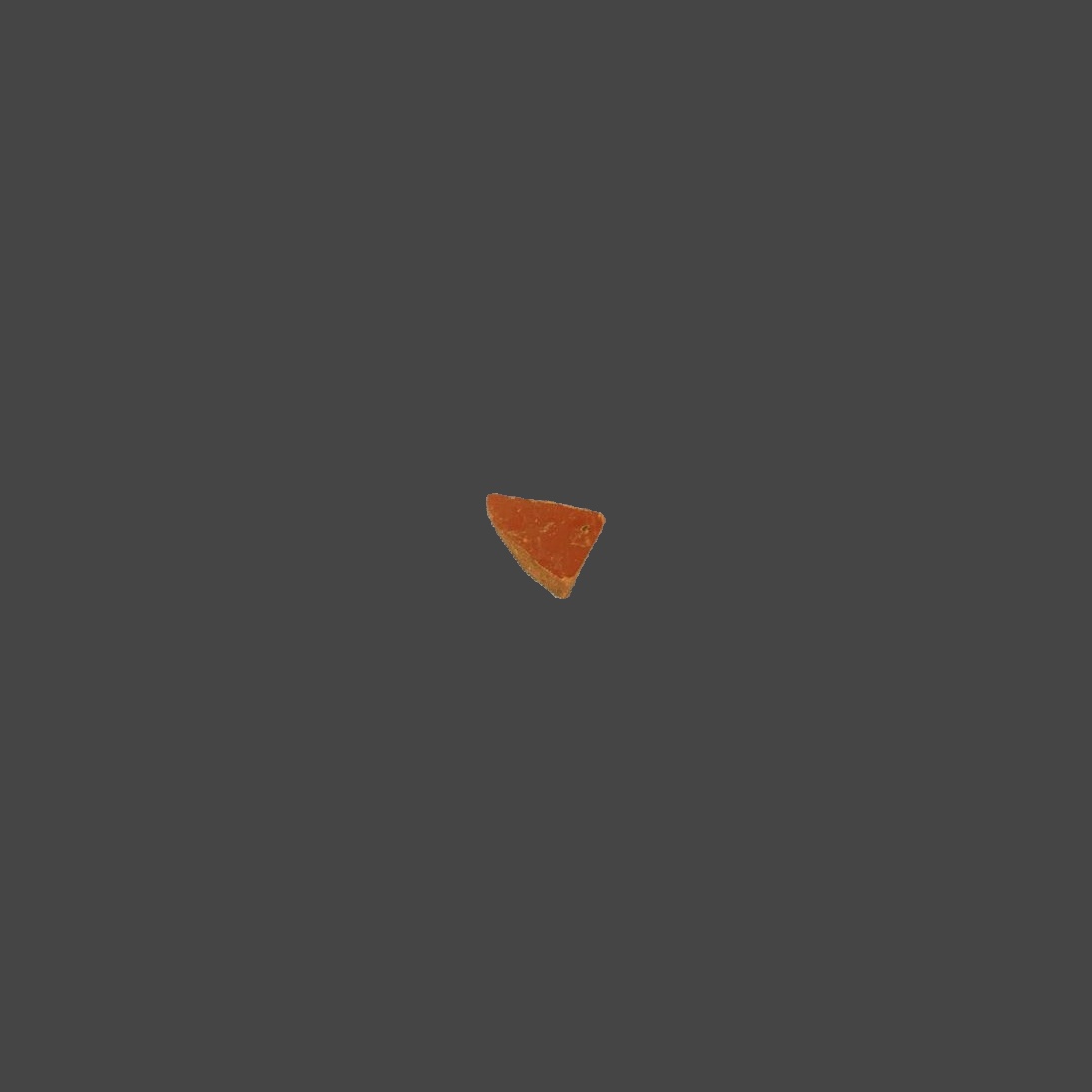}\hspace{1mm}\includegraphics[width=0.15\textwidth]{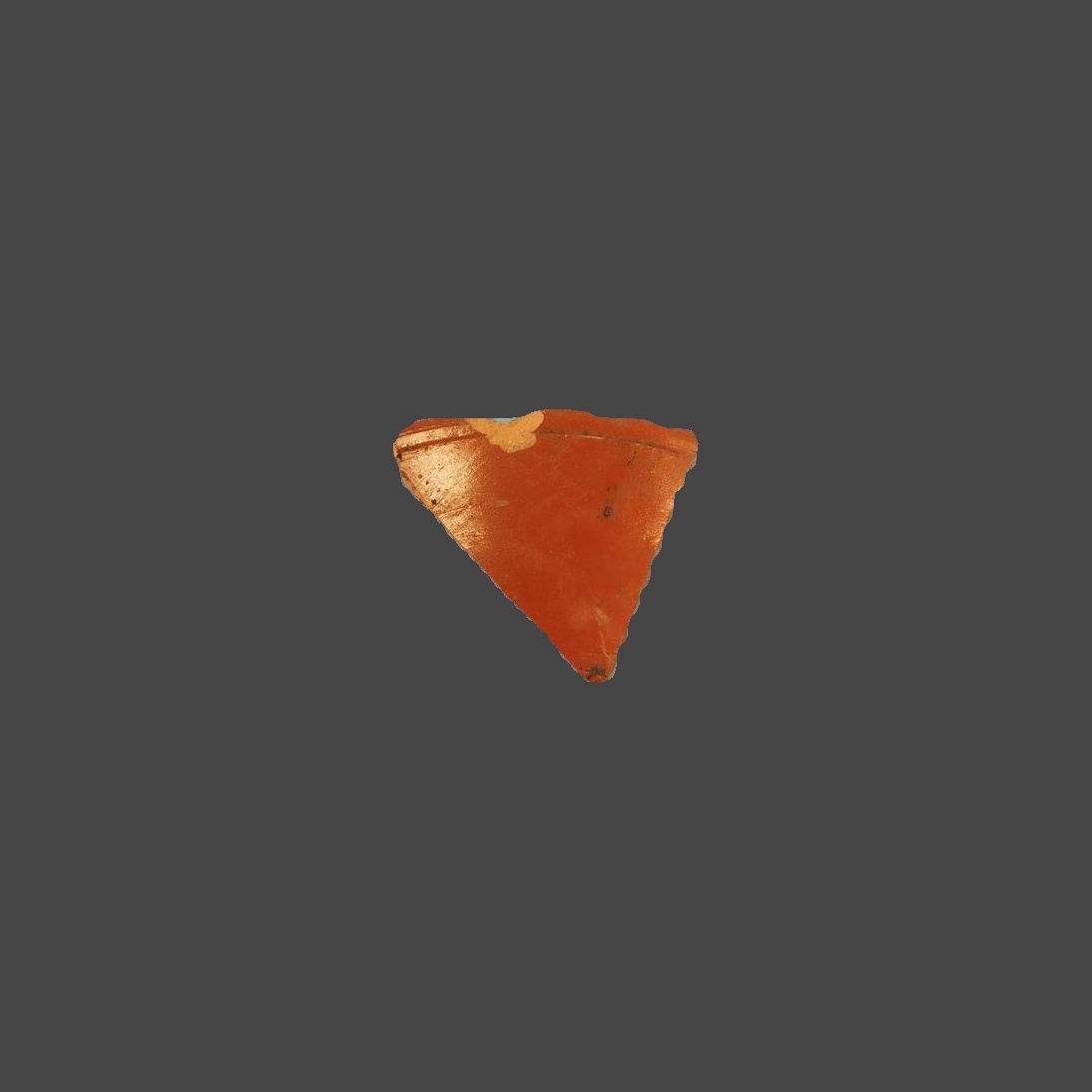}\\
\vspace{2mm}
\includegraphics[width=0.15\textwidth]{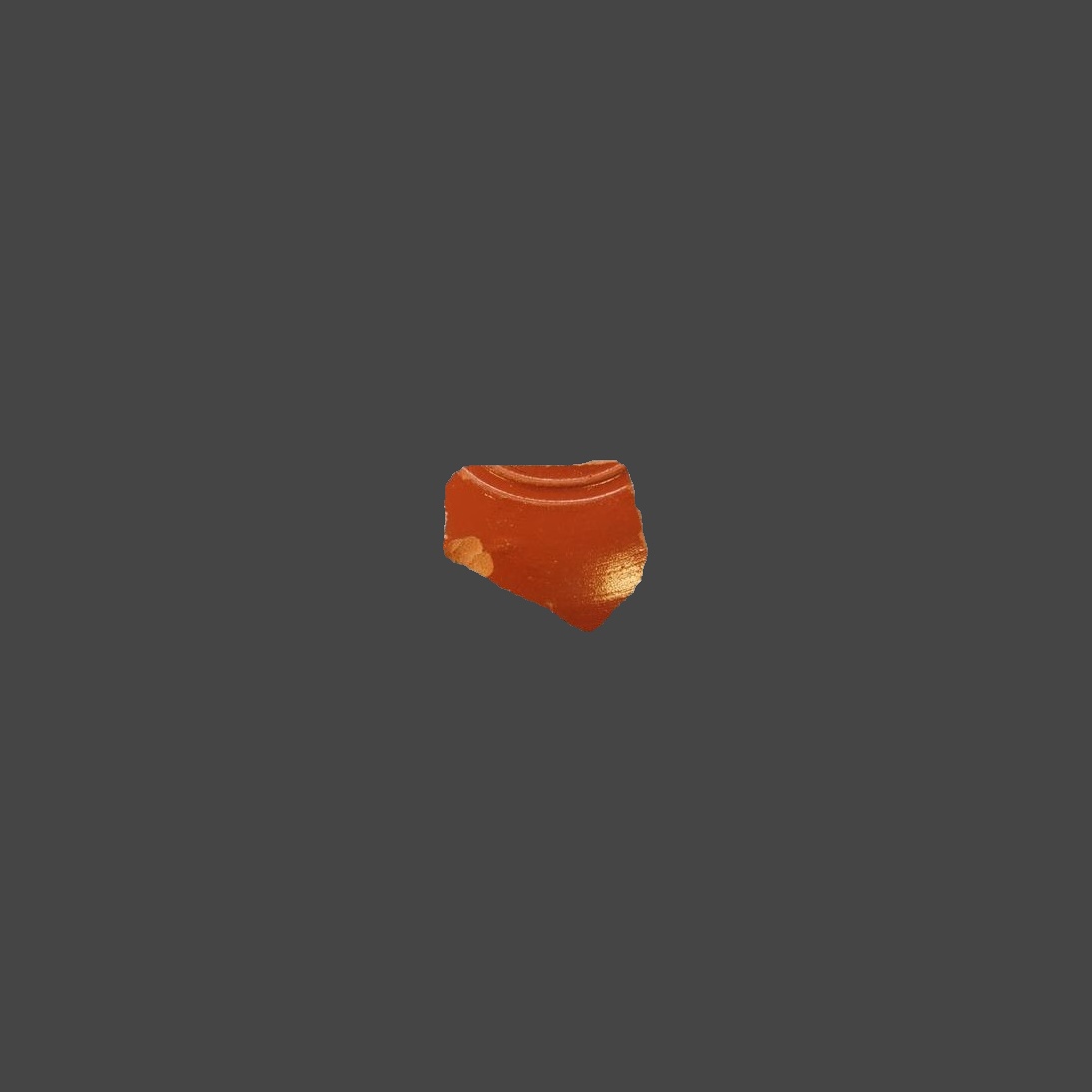}\hspace{1mm}\includegraphics[width=0.15\textwidth]{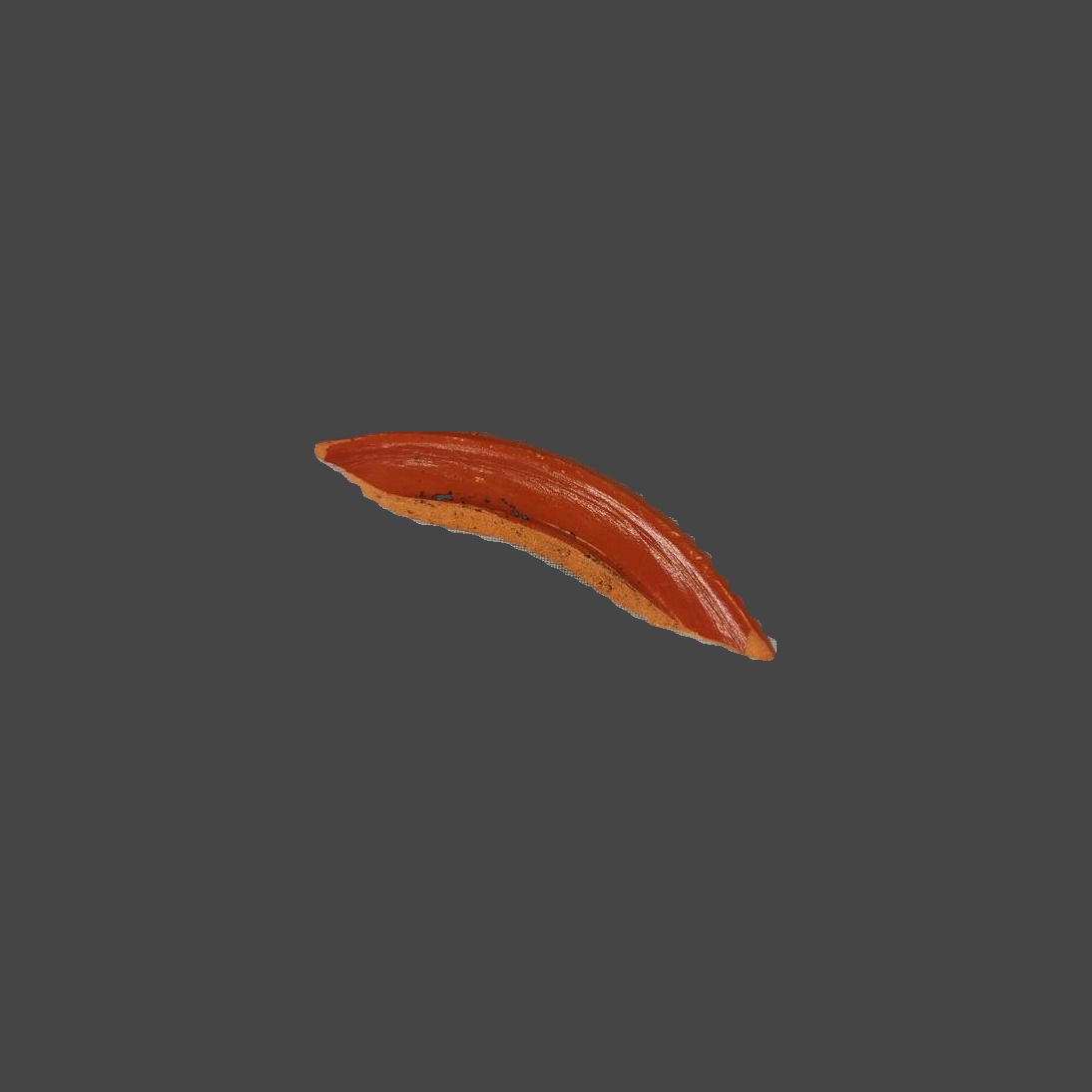}\hspace{1mm}\includegraphics[width=0.15\textwidth]{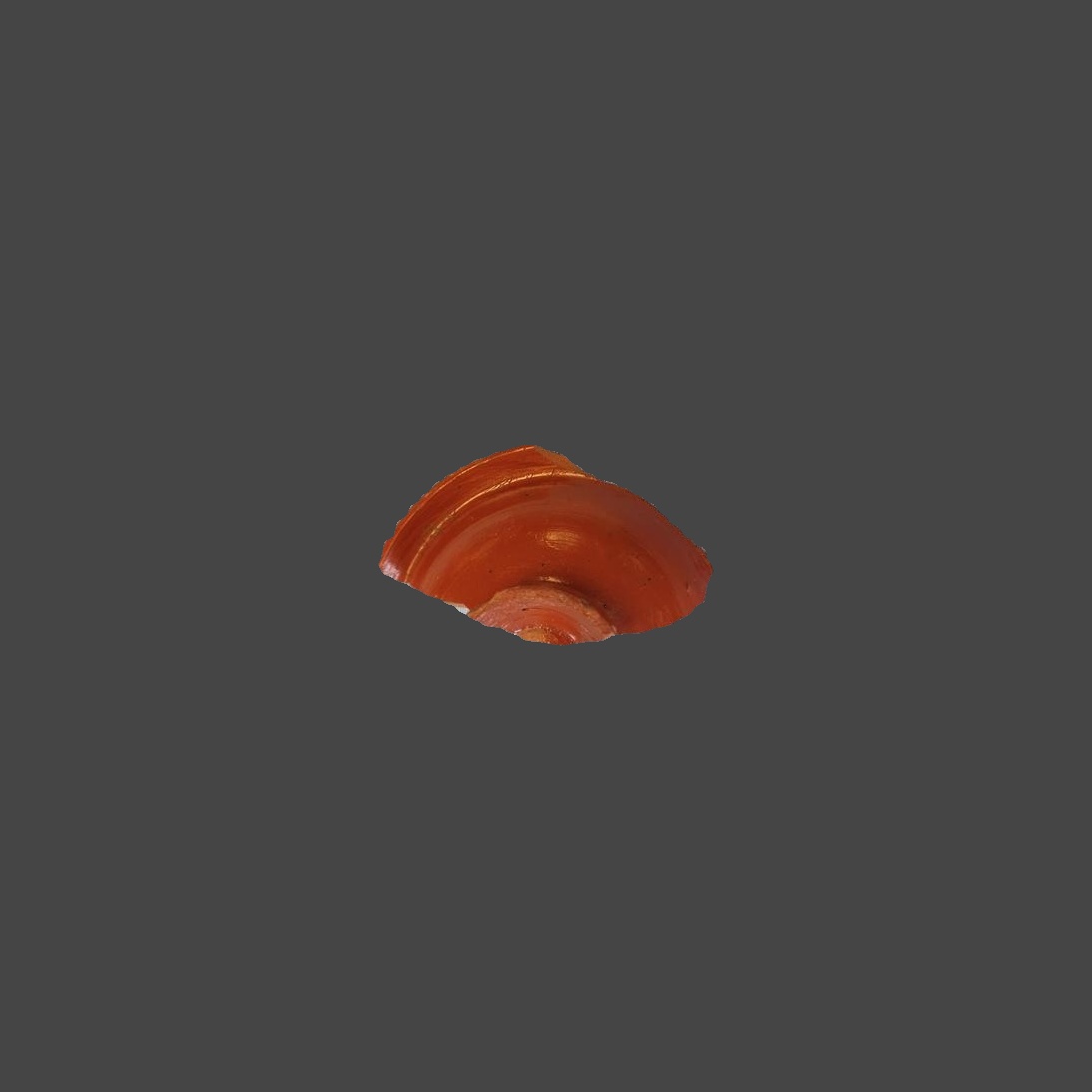}\hspace{1mm}\includegraphics[width=0.15\textwidth]{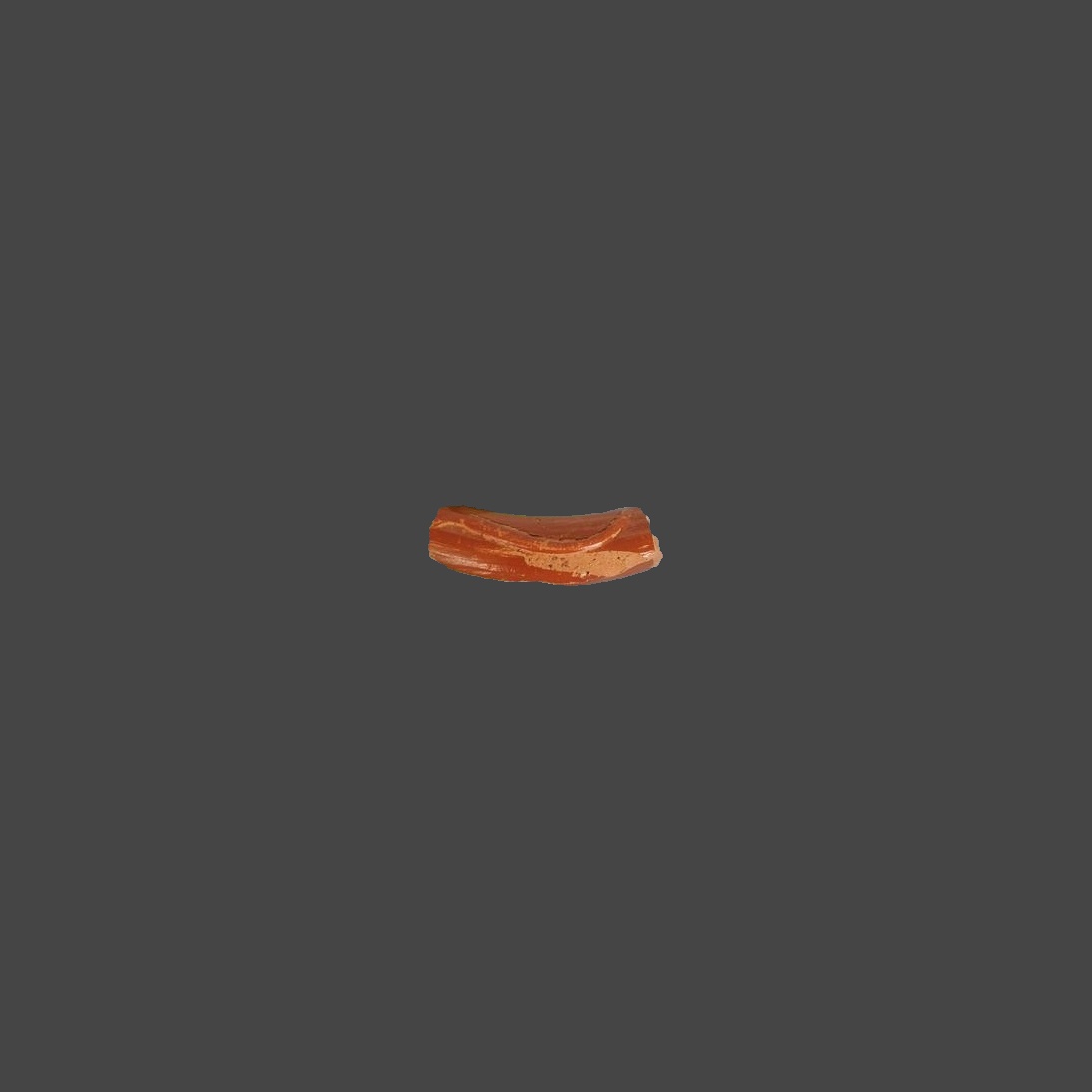}\hspace{1mm}\includegraphics[width=0.15\textwidth]{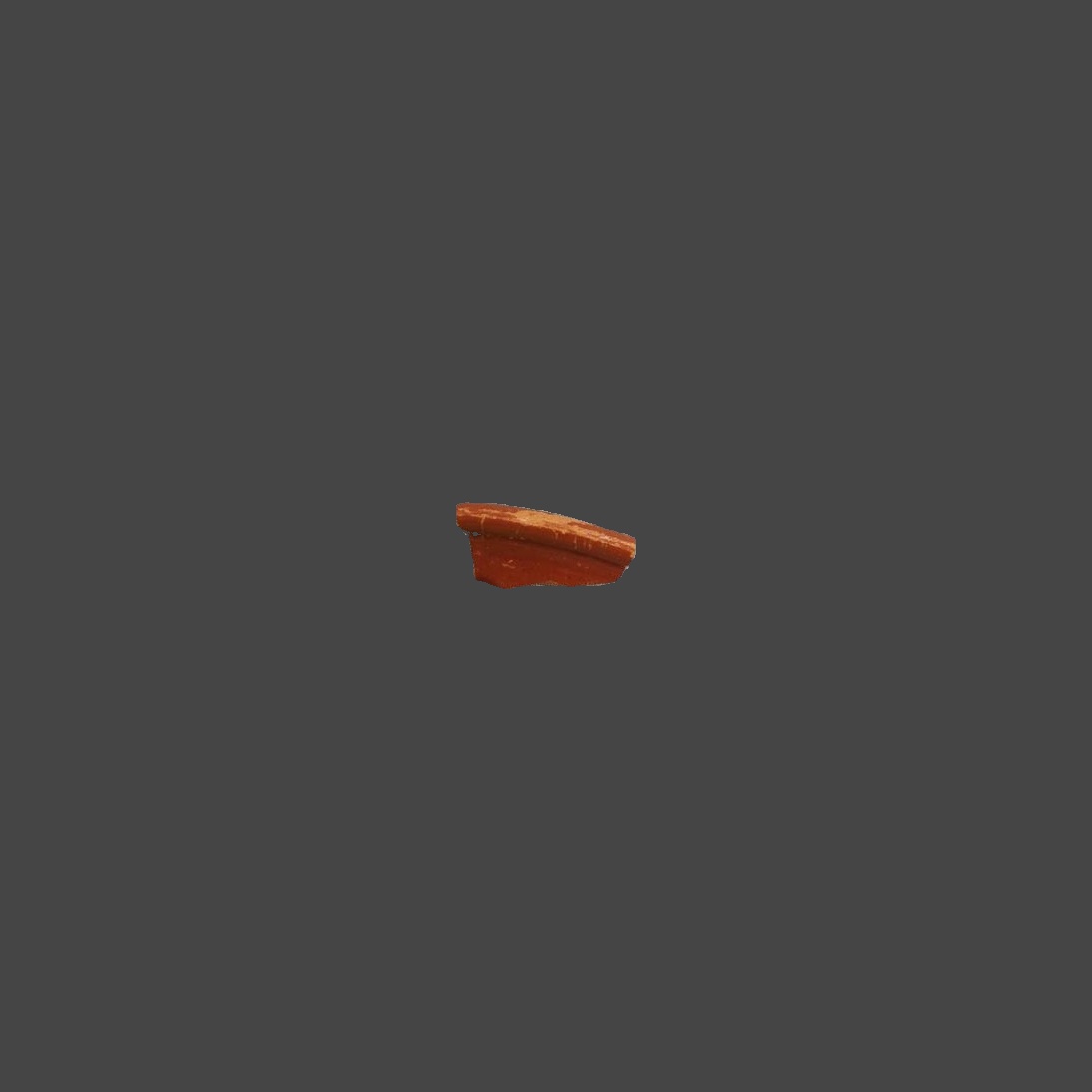}\\
\vspace{2mm}
\includegraphics[width=0.15\textwidth]{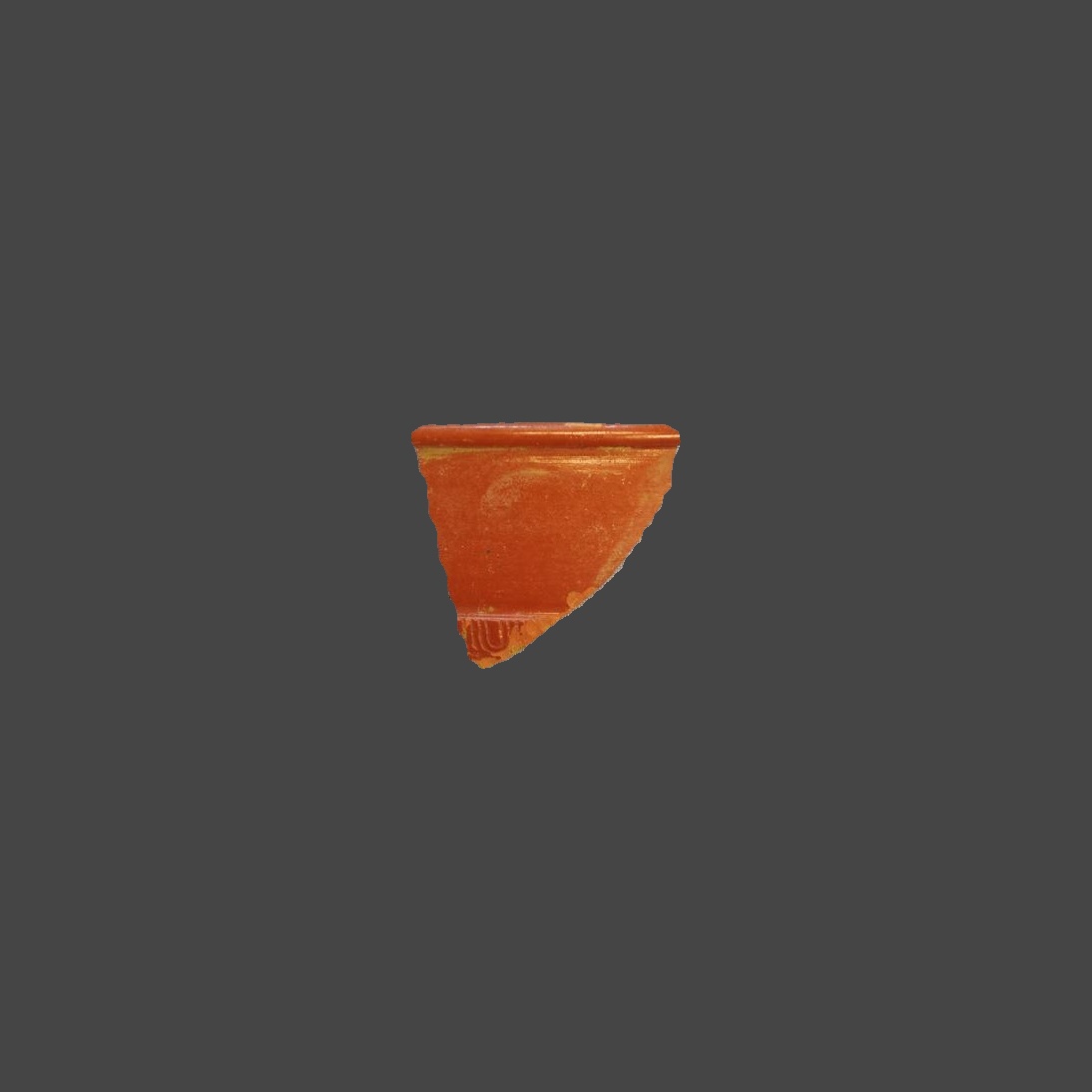}\hspace{1mm}\includegraphics[width=0.15\textwidth]{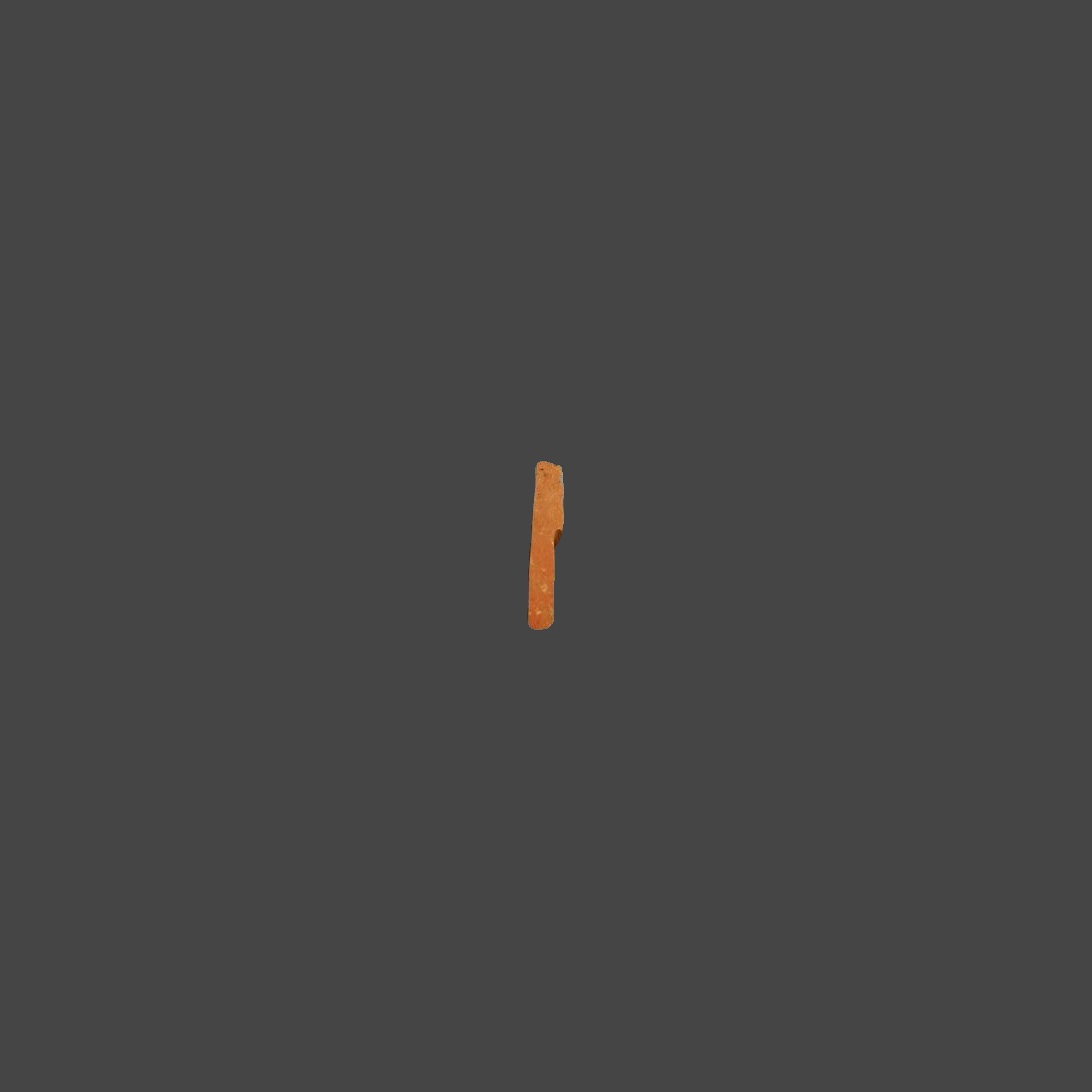}\hspace{1mm}\includegraphics[width=0.15\textwidth]{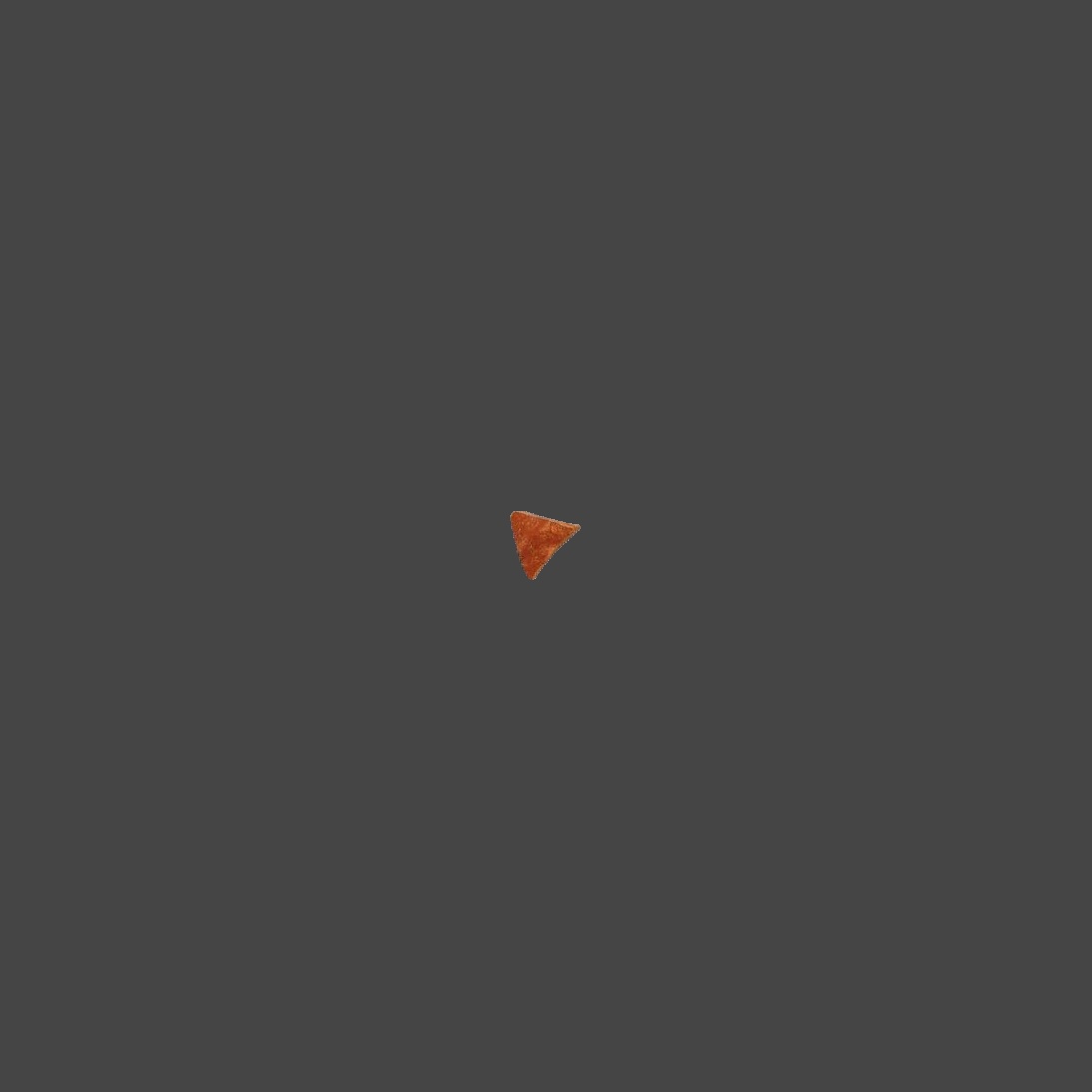}\hspace{1mm}\includegraphics[width=0.15\textwidth]{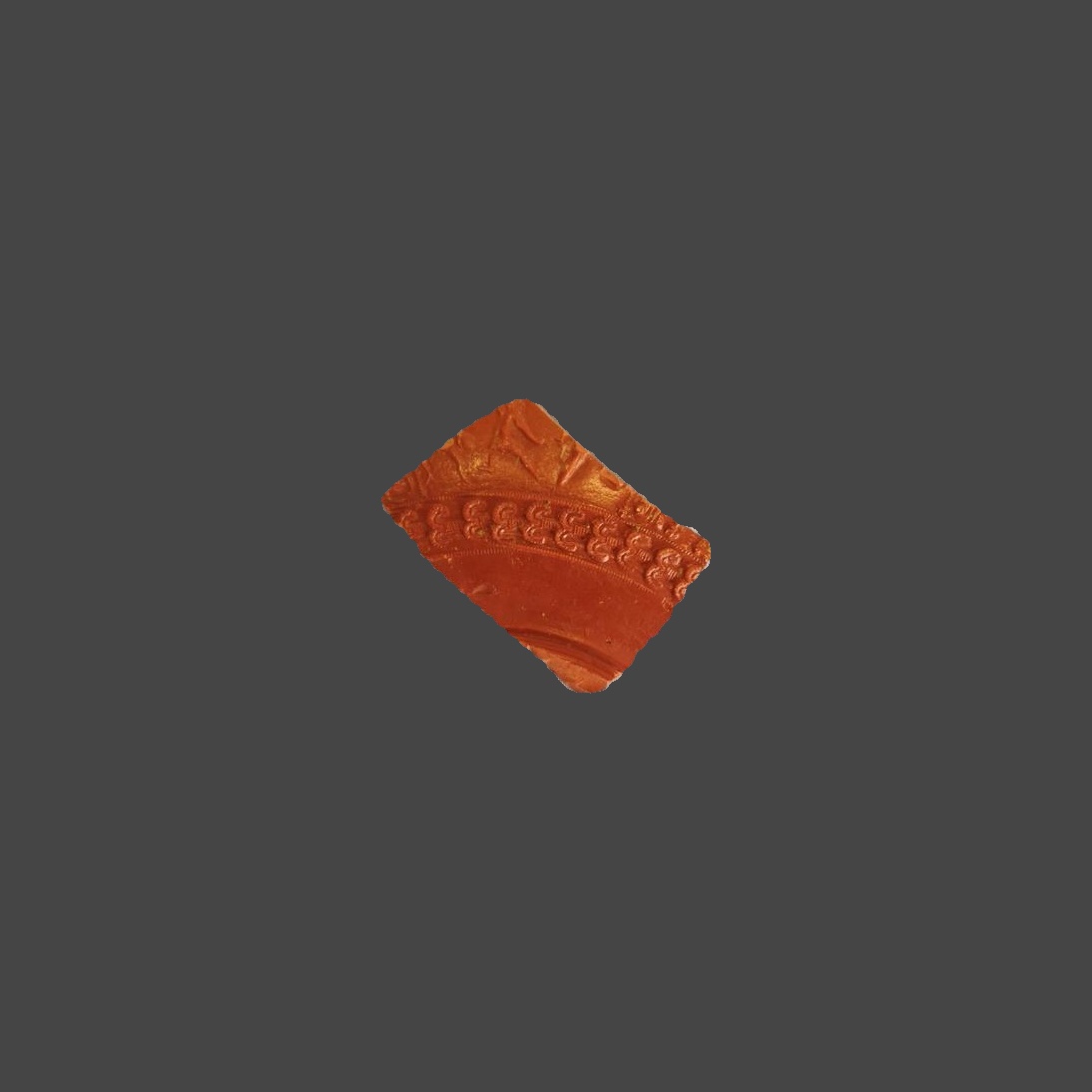}\hspace{1mm}\includegraphics[width=0.15\textwidth]{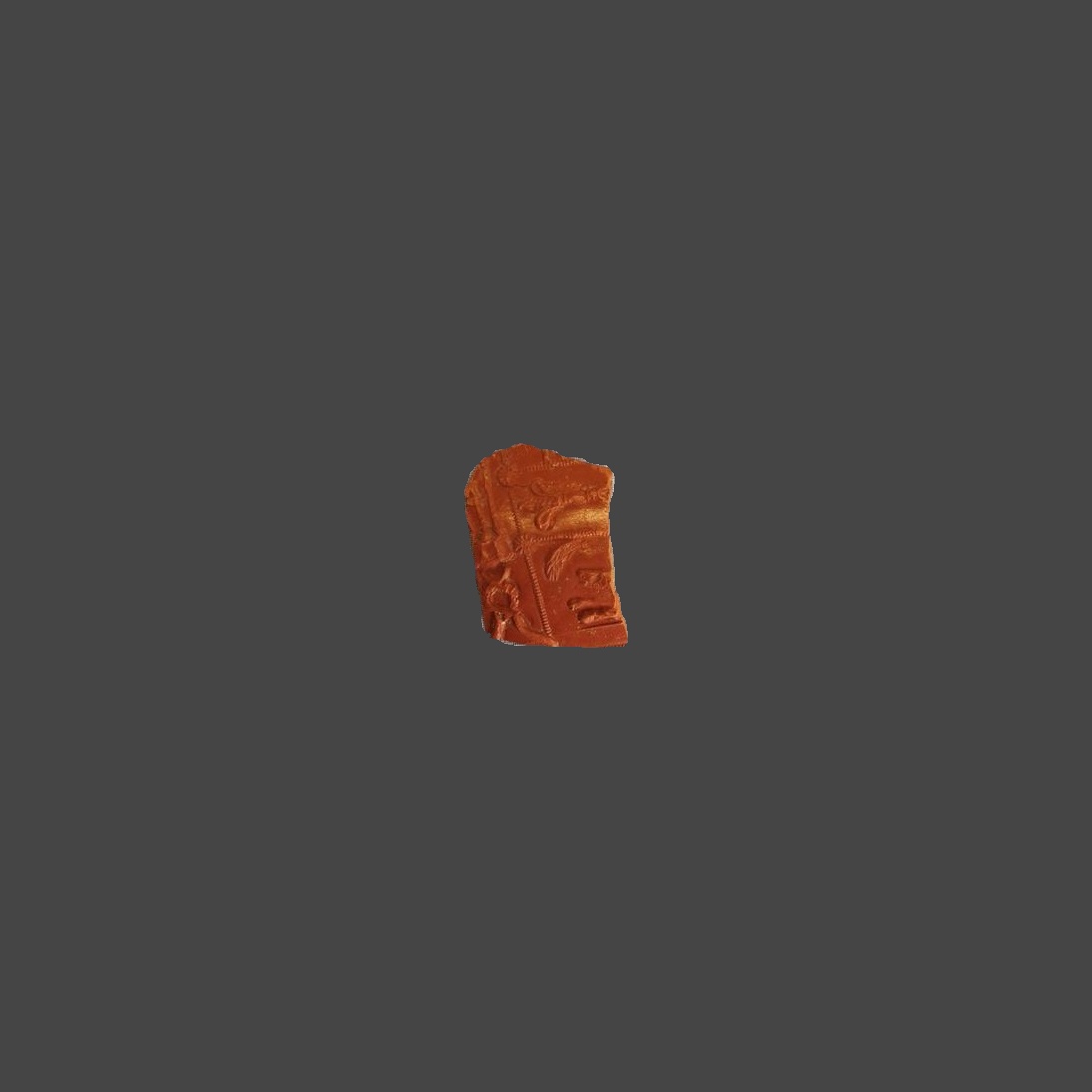}\\
\vspace{2mm}
\includegraphics[width=0.15\textwidth]{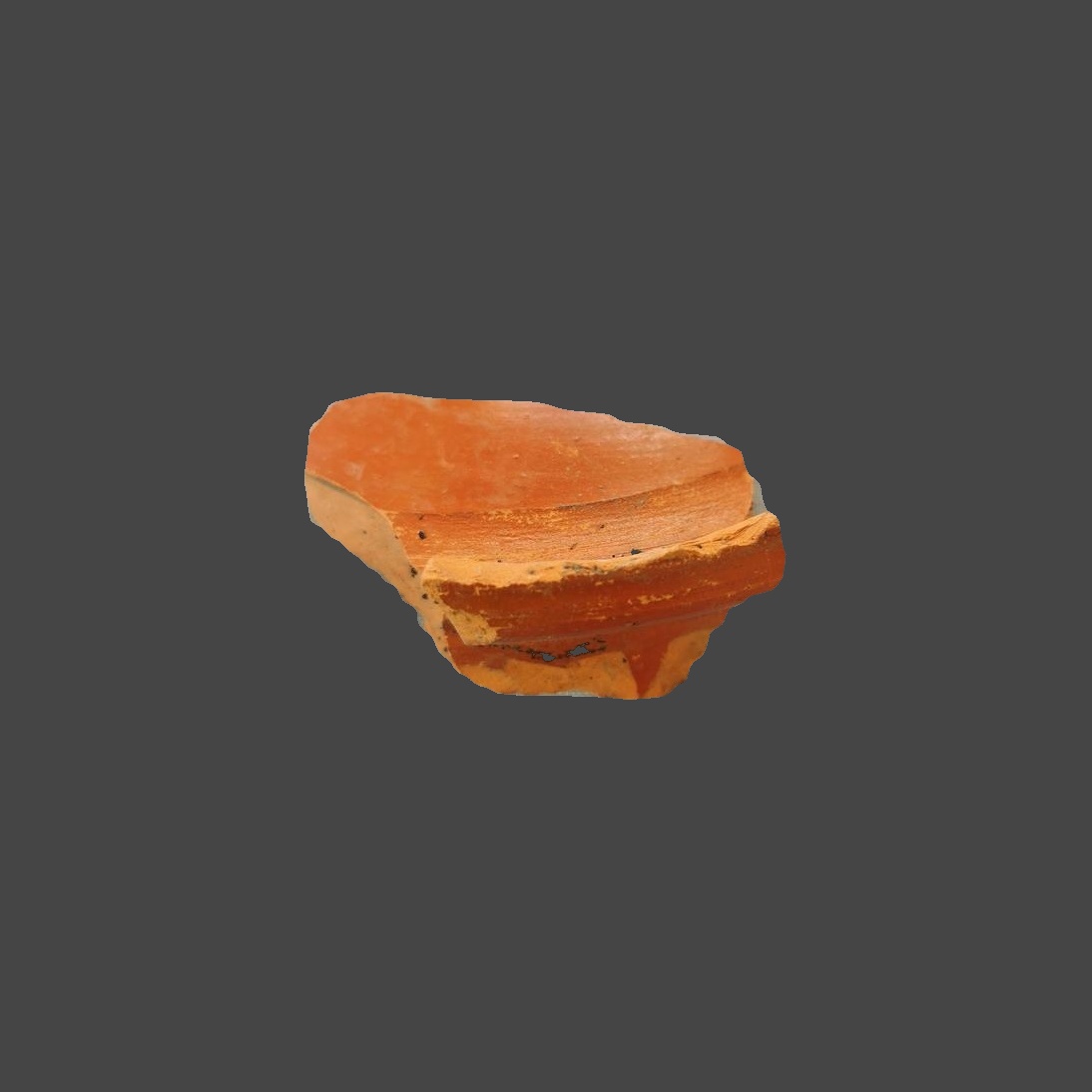}\hspace{1mm}\includegraphics[width=0.15\textwidth]{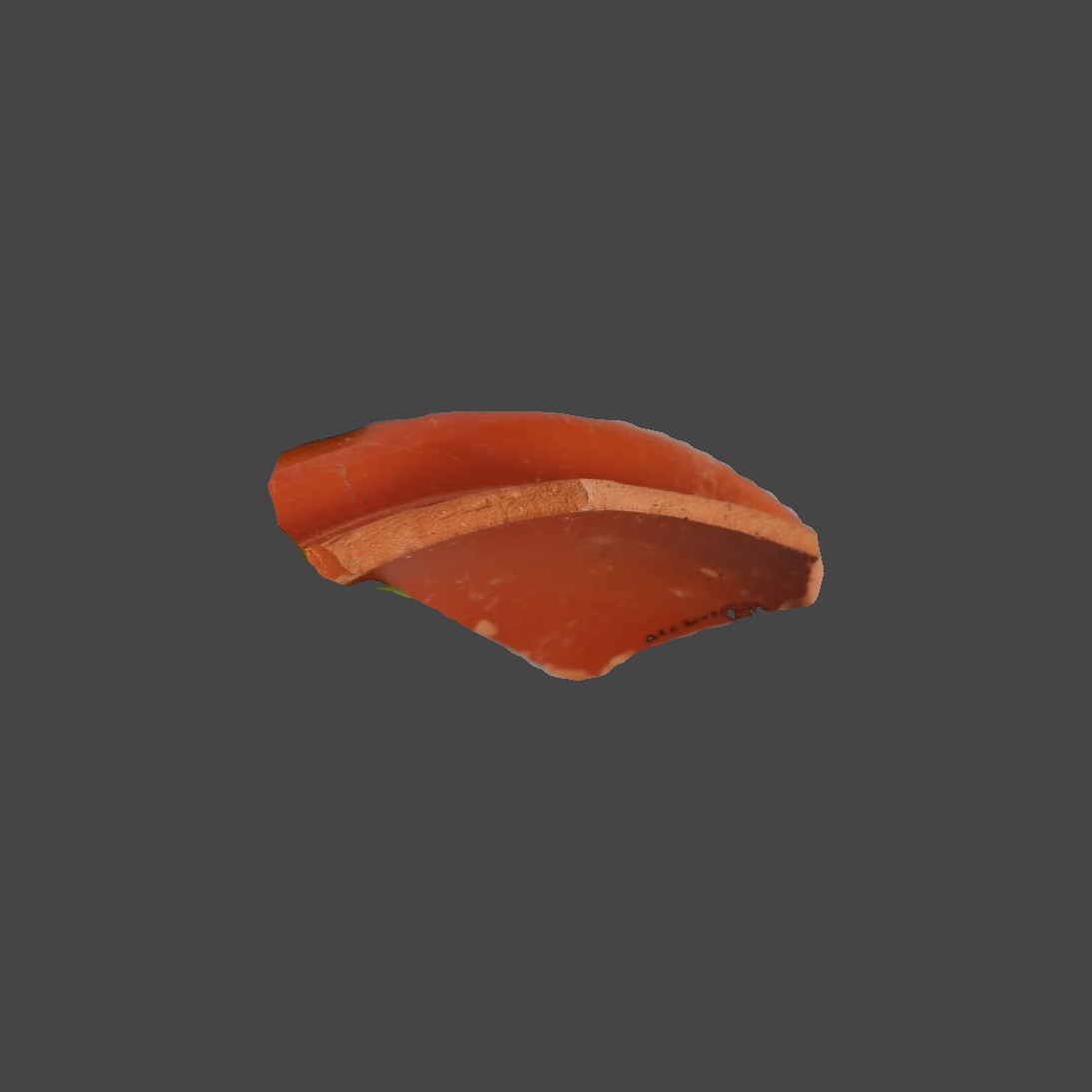}\hspace{1mm}\includegraphics[width=0.15\textwidth]{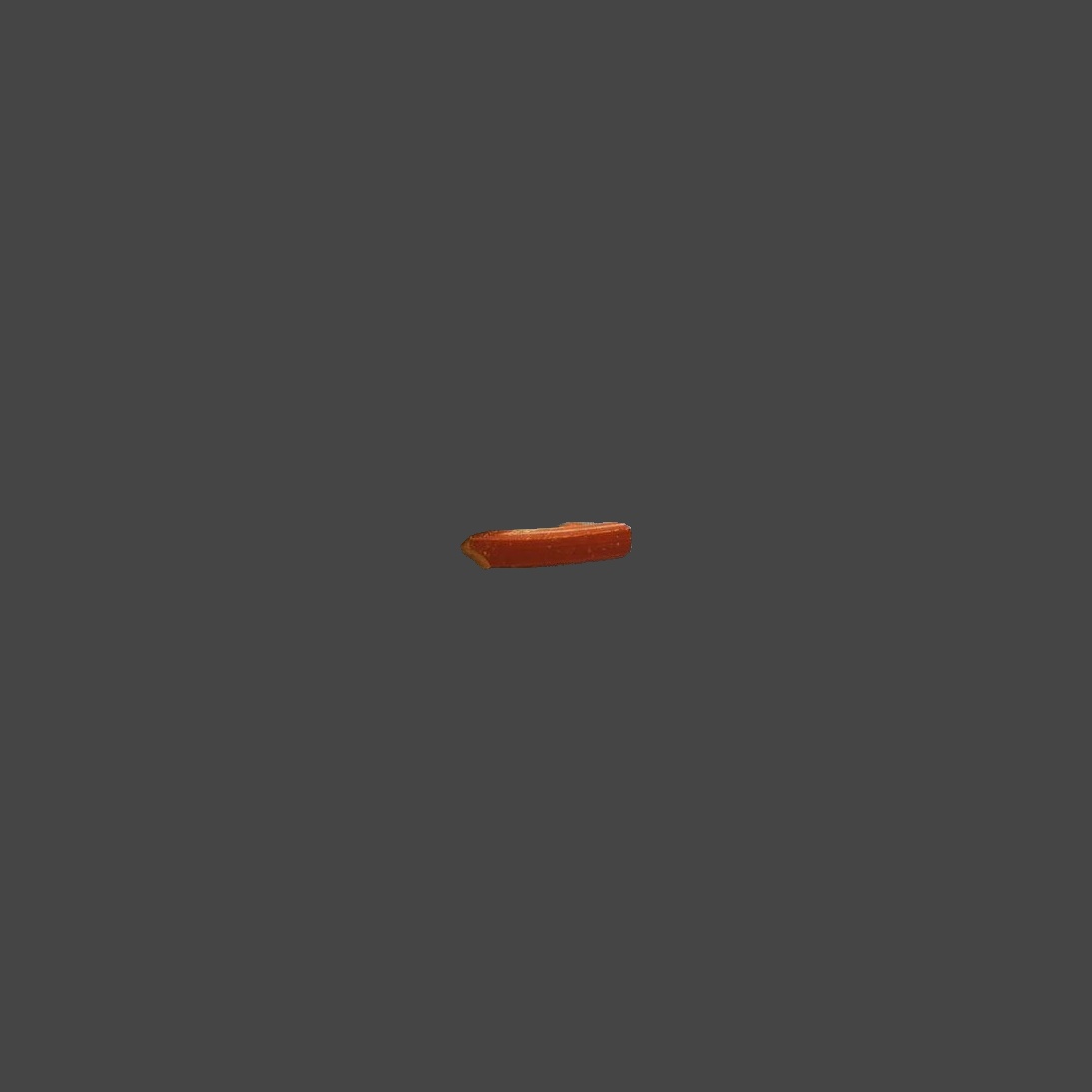}\hspace{1mm}\includegraphics[width=0.15\textwidth]{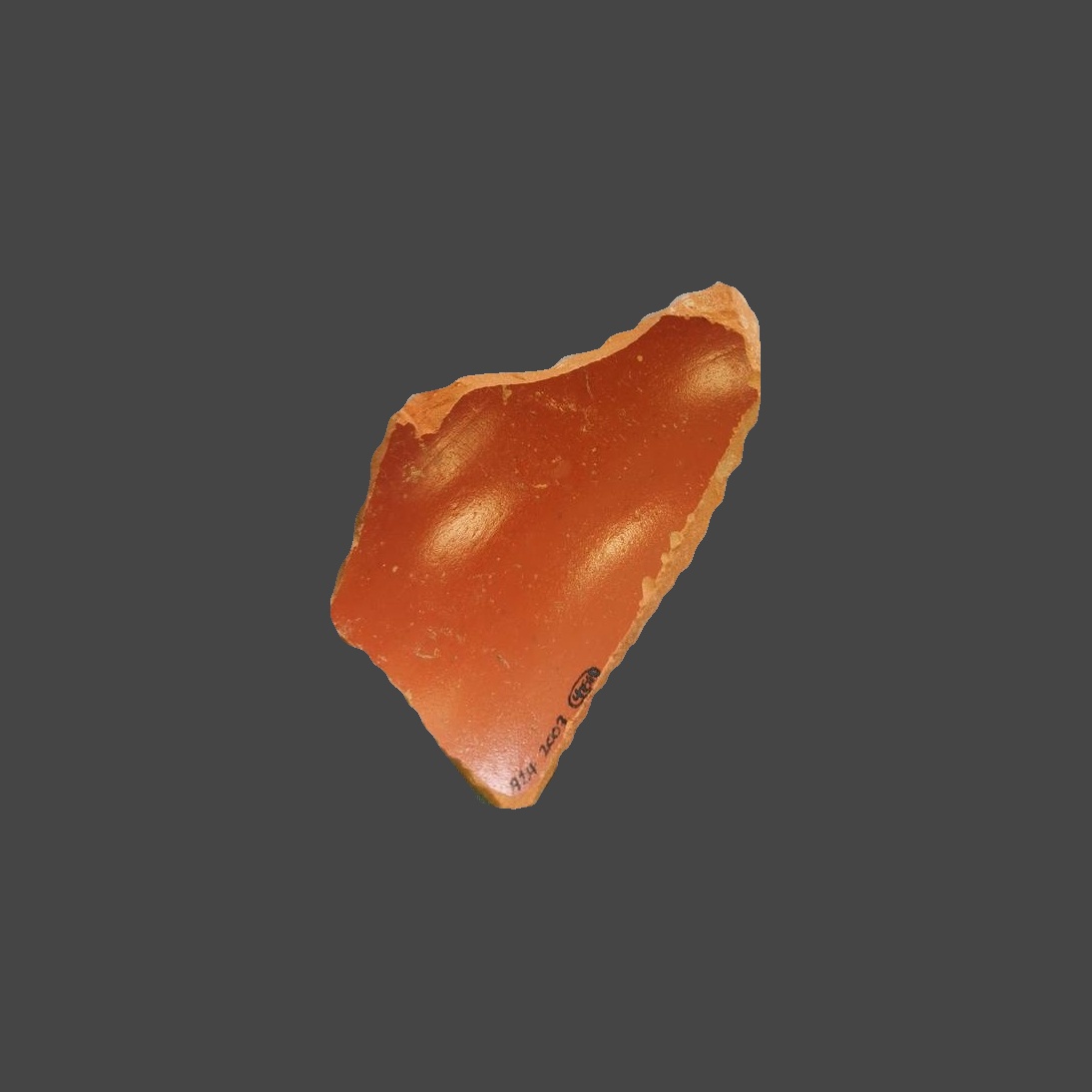}\hspace{1mm}\includegraphics[width=0.15\textwidth]{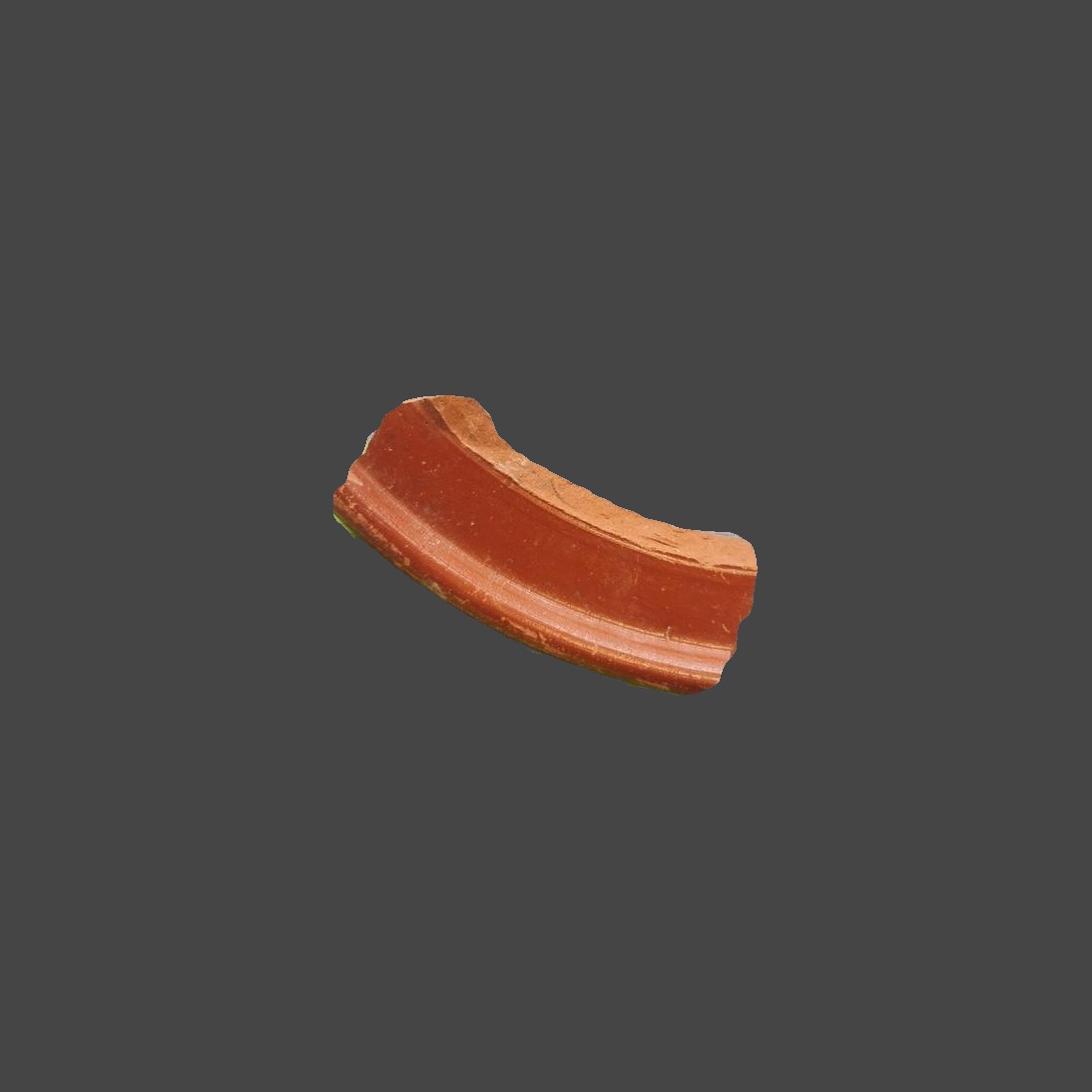}\\

\caption{Examples of images of sherds from different classes. Each row corresponds to a class, with the top row showing images from class $1$, and the bottom row showing images from class $5$.}\label{fig:images_sherds}
\end{figure*}

%
\begin{table}[!t]
\renewcommand{\arraystretch}{1.3}
\caption{Real images available for training, validation, and testing performance of the core pottery classifier}
\label{tab:Training_data_real}
\centering
\begin{tabular}{|c|c|c|c|}
\hline
Class & Training & Validation & Testing\\
\hline
1 & 9858 & 93 & 205\\
2 & 10022 & 97 & 182\\
3 & 639 & 36 & 144 \\
4 & 13994 & 120 & 182 \\
5 & 742 & 84 & 174\\
\hline
\end{tabular}
\end{table}

Direct attempts at training the core classifier with real data consistently produced significantly overfit and under-performing models. We also note a severe class imbalance of the dataset of real images. Therefore, inspired by \cite{nunez2021learning}, we supplemented real data with a synthetic dataset containing $80640$ images of each form training, $10080$ images of each form for validation, and $10080$ images of each form for testing.


\subsubsection{Data for training AI correctors}

To train and test weakly supervised AI correctors, we used additional images of real sherds. The composition of this set is shown in Table \ref{tab:Training_data_corrector}. 

\begin{table}
\caption{Real images available for training and testing performance of the corrector}
\label{tab:Training_data_corrector}
\centering
\begin{tabular}{|c|c|c|}
\hline
Class & Training \\
\hline
1 & 861 \\
2 & 926  \\
3 & 311  \\
4 & 900  \\
5 & 542  \\
\hline
\end{tabular}
\end{table}

\subsection{AI correctors}

In the process of constructing AI correctors we generally followed steps prescribed in Algorithm \ref{alg:corrector_general}. The classifier $F$ was the deep learning InceptionV3 net itself with the modification described above, and the map $\Phi$ was the $512$-dimensional output of the dense layer of the network. The space $\mathcal{H}$ was constrained to the space of all linear mappings from $\Real^d$ to $\Real$. Due to the lack of labeled data, we used $\mathcal{S}$ to construct projectors $h_j$. This is a deviation from the assumption that the choice of $h_j$ is independent of $\mathcal{S}$. In addition, we added an extra step to reduce the dimensionality of vectors $\Phi(u)$. To do so, we determined the principal components $t_1,\dots,t_{d}\in\Real^d$ of the ``positives'' set
\[
\hat{\mathcal{V}}_{+}=\{z \in\Real^d | \ z=\Phi(u), \ (u,y)\in \cup_{j} \mathcal{S}_{+,j}\},
\]
and selected top $k=14$ components which explained $99.87\%$ of the variance of the dataset. 

We also introduce
\[
\begin{split}
&\hat{\mathcal{V}}_{+,j}=\{z \in\Real^d | \ z=\Phi(u), \ (u,y)\in\mathcal{S}_{+,j}\}, \\
&\hat{\mathcal{V}}_{-,j}=\{z \in\Real^d | \ z=\Phi(u), \ (u,y)\in\mathcal{S}_{-,j}\}.
\end{split}
\]

The projectors $h_j$ were then computed as
\begin{equation}\label{eq:Fisher_projectors}
\begin{split}
h_j(x)=&(\Cov(T\hat{\mathcal{V}}_{+,j})+\Cov(T\hat{\mathcal{V}}_{-,j}))^{-1}\times \\
& \times T(\Mean(\hat{\mathcal{V}}_{+,j})-\Mean(\hat{\mathcal{V}}_{-,j})) T x,
\end{split}
\end{equation}
where $T$ is the following $k\times d$ real matrix:
\[ 
T=\left(\begin{array}{c} t_1^{T}\\
                        \cdots\\
                        t_{k}^{T}
                        \end{array}\right).
\]
Note that (\ref{eq:Fisher_projectors}) are the weights of Fisher linear discriminants computed for reduced-dimensional sets $\hat{\mathcal{V}}_{+,j}$, $\hat{\mathcal{V}}_{-,j}$. Our choice of Fisher discriminants over alternatives such as Support Vector Machines is motivated by the numerical and statistical stability of the former provided that the matrices $\Cov(T\hat{\mathcal{V}}_{+,j}), \Cov(T\hat{\mathcal{V}}_{i,j})$ are well conditioned. This allowed to minimise the impact of dependencies of $h_j$ on $\mathcal{S}$. The values of $\Delta_j$, $i=1,...,5$ were as follows $\Delta_1=0.9167$,
$\Delta_2=0.7925$, $\Delta_3=0.8125$, $\Delta_4=0.9622$, $\Delta_5=0.7778$.

\subsection{Training protocol}

The core deep learning neural network was initially trained in Tensorflow 2.15 \& Keras, Python 3.11, on simulated data for $10$ epochs using ADAM with the initial learning rate equal to $0.0001$. The size of mini-batches was set to $32$. Data was augmented with the following transformations: random zoom up to $0.3$, random vertical flips, and random rotation from $0$ to $180$ degrees. After this initial phase, images of simulated data were removed from the training set, and images of real sherds were added instead (see Table \ref{tab:Training_data_real}). The network was trained for further $15$ epochs. At this stage, the dynamics of training and validation accuracy started to diverge indicating the onset of potential overfitting. 
  
\subsection{Results and discussion}

Performance of the trained core model on the images of real sherds used in the model's training is shown in Table \ref{tab:results_net_training_accuracy}.

\begin{table}[]
\caption{Performance of the core classifier on training data}
\label{tab:results_net_training_accuracy}
\centering
\begin{tabular}{|c|c|c|c|c|c||c|}
\hline
Class & 1 & 2 & 3 & 4 & 5 & Recall\\
\hline
1 & 4330  & 284 &  102 & 4942  & 200 & 0.44\\
2 &  455  & 1282 &   30 & 8116 &  139 & 0.13\\
3 & 130  &  10  & 146  & 329    & 24 & 0.23\\
4 & 101  &  81  &  11 & 13763   & 38 & 0.98\\
5 & 36  &  11   &  8  & 467  & 220 & 0.3\\
\hline
\end{tabular}
\end{table}

The values of recall on the test set ($0.64$, $0.08$, $0.11$, $0.95$, $0.21$) were consistent with those for the training data but were noticeably worse overall.  

\begin{table}[]
\caption{Performance of corrected model on training data (Table \ref{tab:Training_data_corrector})}
\label{tab:results_corrected_training_accuracy}
\centering
\begin{tabular}{|c|c|c|c|c|c|}
\hline
Class & 1 & 2 & 3 & 4 & 5 \\
\hline
Correct & 145  & 78 &  26 & 182  & 75 \\
Incorrect &  28  & 19 &  21 & 61 &  18\\
Rejected & 536 & 63 &  51 & 2189    & 48\\
\hline
Proportion of&   &    &   &    &  \\
accepted ``incorrect'' data &  0.09 & 0.36 & 0.33 & 0.039 & 0.4\\
\hline
Theoretical  upper&  &  &    &     &  \\
bound: 1- $\rho(\Delta_j,M_{-,j})$ & 0.19   &   0.43   &  0.39 & 0.09   & 0.46\\
\hline
Conditional Recall & 0.84 & 0.8 & 0.55 & 0.75 &  0.81\\
\hline
\end{tabular}
\end{table}

\begin{table}[]
\caption{Performance of corrected model on test data}
\label{tab:results_corrected_testing_accuracy}
\centering
\begin{tabular}{|c|c|c|c|c|c|}
\hline
Class & 1 & 2 & 3 & 4 & 5 \\
\hline
Correct & 38  & 11 &  13 & 15  & 31 \\
Incorrect &  9  & 5 &  7 & 17 &  11\\
Rejected & 160 & 15 &  8 & 537    & 10\\
\hline
Proportion of &   &    &   &    &  \\
accepted ``incorrect'' data &  0.12 & 0.31 & 0.58 & 0.043 & 0.73\\
\hline
Conditional Recall & 0.81 & 0.69 & 0.65 & 0.47 &  0.74 \\
\hline

\end{tabular}
\end{table}

The performance of the system with weakly supervised AI correctors is shown in Tables \ref{tab:results_corrected_training_accuracy} and \ref{tab:results_corrected_testing_accuracy}. According to these tables, significant improvements in recall conditioned on accepted answers (Conditional Recall in the tables), for Classes $1$, $2$, $3$, and $5$ are observed. At the same time, we see that the value of condition recall for Class $4$ dropped relative to what we have seen for the core classifier (see Table \ref{tab:results_net_training_accuracy}). This, however, should not be a concern as excellent recall for a single class does not necessarily equate to excellent accuracy. A simple example is a classifier always returning the same class label no matter what its inputs are.

In Table  \ref{tab:results_corrected_training_accuracy} we also show theoretical upper bounds $1-\rho(\Delta_j,M_{-,j})$ on the proportion of errors which the system with correctors has mistakenly classified as correct (for each class). For the correctors' training dataset, the values of $M_{-,j}$ were: $M_{-,1}=288$, $M_{-,2}=53$, $M_{-,3}=64$, $M_{-,4}=1562$, $M_{-,5}=45$. As we see from Table  \ref{tab:results_corrected_training_accuracy}, these figures are in agreement with what was observed in experiments. We note that the empirically estimated proportion of the same quantities on the test set for Class 3 and Class 5 exceeded the theoretical bound. This deviation, however, was likely because of the cardinalities of the sets of errors for these classes were small: $M_{-,3}=12$ and $M_{-,5}=15$.

Overall, we see that the application of weakly supervised error correctors improves the performance of the classifier on the data for which the corrected model is confident to produce an output. Moreover, for these answers one can immediately generate bounds on the probabilities of errors (see Table \ref{tab:results_corrected_training_accuracy}).

\section{Conclusion}\label{sec:Conclusion}

In this work, we presented a novel methodology for dealing with AI errors. The methodology builds on two fundamental ideas: the idea of error correctors (see e.g. \cite{gorban2018correction,gorban2021high,gorban2018blessing}) and the idea of the abstaining option \cite{abstaining_2021,bartlett2008classification,chow1970optimum}. Here we used these ideas in a general setting when the original classifier's state, which may potentially be high or even infinite-dimensional, could be projected onto some low-dimensional subspace. We used this ``projected'' information to build weakly-supervised correctors to moderate the decisions of the original classifier. Projection onto appropriate low-dimensional subspace enables computing rigorous bounds on the probabilities of correct decisions conditional upon the classifier's outputs. These bounds are distribution-agnostic in the sense that no knowledge of the distribution is needed to produce such bounds. The theory is illustrated with an example of a real-world hard and significantly under-sampled problem of the classification of archaeological artefacts. 

The machinery involved in constructing correctors allows them to be essentially weakly supervised learners as the bounds we derived become practically meaningful for low- and moderately- labelled datasets (when mere dozens or hundreds of labeled errors can be made available to the designer or human supervisor). This opens a pathway for the application of results to a broad class of settings with strong but expensive AI models moderated by weakly supervised and computationally cheap correctors. Another interesting question could be in exploring conditioning on more complicated patterns of the classifier's outputs. This, however, falls outside of the scope of our current contribution and will be considered in future work.






\bibliographystyle{IEEEtran}
\bibliography{IEEEabrv,references}

\end{document}